\documentclass[journal]{IEEEtran}
\usepackage{cite, color}
\usepackage[pdftex]{graphicx}
\usepackage[export]{adjustbox}
\usepackage{subcaption}

\usepackage{bm}
\usepackage{amssymb,amsmath}
\usepackage{amsthm}

\newtheorem{definition}{Definition}

\title{c-Eval: A Unified Metric to Evaluate Feature-based Explanations via Perturbation}

\author{\normalsize
	\IEEEauthorblockN{Minh N. Vu\IEEEauthorrefmark{1}, Truc D. Nguyen\IEEEauthorrefmark{1}, NhatHai Phan\IEEEauthorrefmark{2}, Ralucca Gera\IEEEauthorrefmark{3}, and My T. Thai\IEEEauthorrefmark{1}}\\
	\IEEEauthorblockA{
		\IEEEauthorrefmark{1} University of Florida,
	    Gainesville, Florida, USA}
	\IEEEauthorblockA{
		\IEEEauthorrefmark{2} New Jersey Institute of Technology, Newark, New Jersey, USA}
	 \IEEEauthorblockA{
		\IEEEauthorrefmark{3} Naval Postgraduate School,
	    Monterey, California, USA}
}
 
\begin{document}

\maketitle

\begin{abstract}

In many modern image-classification applications, understanding the cause of model's prediction can be as critical as the prediction's accuracy itself. Various feature-based local explanations generation methods have been designed to give us more insights on the decision of complex classifiers. Nevertheless, there is no consensus on evaluating the quality of different explanations. In response to this lack of comprehensive evaluation, we introduce the c-Eval metric and its corresponding framework to quantify the feature-based local explanation's quality. Given a classifier's prediction and the corresponding explanation on that prediction, c-Eval is the minimum-distortion perturbation that successfully alters the prediction while keeping the explanation's features unchanged. We then demonstrate how c-Eval can be computed using some modifications on existing adversarial generation libraries. To show that c-Eval captures the importance of input's features, we establish the connection between c-Eval and the features returned by explainers in affine and nearly-affine classifiers. We then introduce the c-Eval plot, which not only displays a strong connection between c-Eval and explainers' quality, but also helps automatically determine explainer's parameters. Since the generation of c-Eval relies on adversarial generation, we provide a demo of c-Eval on adversarial-robust models and show that the metric is applicable in those models. Finally, extensive experiments of explainers on different datasets are conducted to support the adoption of c-Eval in evaluating explainers' performance.
\end{abstract}


\IEEEpeerreviewmaketitle

\section{Introduction}

With the pervasiveness of machine learning in many emerging domains, especially in critical applications such as health-care or autonomous systems, it is utmost important to understand why a machine learning model makes such a prediction. For example, deep convolutional neural networks have been able to classify skin cancer at a level of competence comparable to dermatologists~\cite{Esteva2017}. However, doctors cannot act upon these predictions blindly. Providing additional intelligible explanations such as a highlighted skin region that contributes to the prediction will aid doctors significantly in making their diagnoses. Along this direction, many machine learning explainers supporting users in interpreting the predictions of complex neural networks on given inputs, called local explainers, 
have been proposed and studied, such as SHAP~\cite{Scott2017}, LIME~\cite{Marco2016}, Grad-CAM (GCam)~\cite{Ramprasaath2016}, and DeepLIFT~\cite{Avanti2017},  among others~\cite{Bach2015,Springenberg2014,Simonyan2013,Daniel2017,Mukund2017,Robnik2008,Strumbelj2009,Martens2014}. Since the outputs of these explainers are subsets of input features or weights on the input's features, these explainers are referred as feature-based local explainers. 

\begin{figure}[ht]
	\centering
% 	\hspace{5mm}
% 	\vspace{2mm}
	\begin{subfigure}{.112\textwidth}
		\centering
		\includegraphics[scale=0.25]{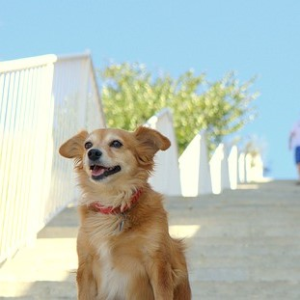}
		\caption{Original}
		\label{fig:sfig1}
	\end{subfigure}
	\begin{subfigure}{.112\textwidth}
		\centering
		\includegraphics[scale=0.25]{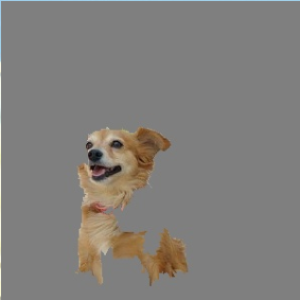}
		\caption{LIME}
		\label{fig:sfig2}
	\end{subfigure}
	\begin{subfigure}{.112\textwidth}
		\centering
		\includegraphics[scale=0.25]{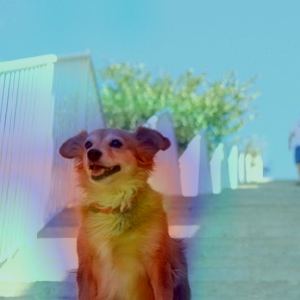}
		\caption{GCam}
		\label{fig:sfig3}
	\end{subfigure}
	\begin{subfigure}{.112\textwidth}
		\centering
		\includegraphics[scale=0.25]{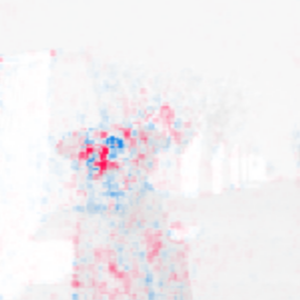}
		\caption{SHAP}
		\label{fig:sfig4}
	\end{subfigure}
	\caption{Explanations generated by different feature-based local explainers of the prediction \textit{Pembroke} of Inception-v3 image classifier.}
	\label{difex}
% 	\vspace{-2mm}
\end{figure}

As there exist many feature-based local explainers, it is important to evaluate the quality of their outputs, called explanations. Unfortunately, evaluating explanations remains as a daunting task \cite{Lipton2016IML,Kim2015}. One major challenge in evaluating explanations is the lack of ground-truth explanations, i.e. many neural networks remain black-box. In fact, most feature-based explanations have been evaluated only through a small set of human-based experiments which apparently does not imply the global guarantee on their quality \cite{Scott2017,Avanti2017}. Another challenge is a diverse presentation of different explanations. Fig.~\ref{difex} shows an example of three explanations generated by LIME, GCam, and SHAP explainers for the prediction \textit{Pembroke} made by the Inception-v3 image classifier~\cite{Inception2015}. All of them highlight the region containing the \textit{Pembroke}; however, their formats vary from picture segments in LIME, heat-map in GCam to pixel importance-weights in SHAP. Furthermore, explainers might be designed for different objectives as there is a fundamental trade-off between the interpretability and the accuracy of explanations~\cite{Marco2016,Scott2017}. In fact, one explanation can be utilized for end-users to interpret, but its consistency with the explained prediction might be lost. The diversity in presentations and objectives constitutes a great challenge in evaluating different explanations. 

{\bf Contribution.} In this research, we focus on evaluating explanations of feature-based local explainers which are used to interpret individual prediction of black box  models. % In order to assess the quality of an explainer, we first need to evaluate the quality of explanations which are generated by the explainer. Along this direction, therefore,
We first introduce a novel metric, $c$-Eval, to evaluate the quality of explanations. We exploit an intuition that certain features are important to the model's prediction only if it is difficult to change the prediction when those features are kept intact. The quality of an explanation is therefore quantified by the minimum amount of perturbation on features outside of the explanation that can alter the prediction. We further provide analysis showing the connection between the importance of features containing in an explanation and its corresponding $c$-Eval in multi-class affine classifiers. For general non-affine classifiers, our experimental results based on $c$-Eval suggests the existence of nearly-affine decision surfaces in many modern classifiers. 
This observation encourages an adoption of $c$-Eval metric in evaluating explanations of predictions made by a broad range of image classifiers.
Additionally, we introduce the $c$-Eval plot, an approach based on $c$-Eval to visualize explainers' behaviors on a given input. Using LIME explainer as an example, we show how $c$-Eval plot helps us gain more trust on LIME and select appropriate parameters for the explainers. We also heuristically demonstrate the behaviors of $c$-Eval in adversarial-robust models. Our results show that the $c$-Eval computed in robust modes is highly correlated with the non-robust counterpart, which strengthens and validates the applications of $c$-Eval in robust models. Finally, extensive experiments are conducted for various explainers %on different models 
to support the usage of $c$-Eval in evaluating explanations. 

% Our contributions in this research can be summarized as follows:
% \begin{enumerate}
% 	\item We introduce $c$-Eval metric to evaluate the quality of explanations generated by any feature-based local explainers.
% 	\item We provide the analysis connecting the $c$-Eval and the importance of input features in affine classifiers: an explanation with higher $c$-Eval implies the features included in the explanation are more aligned with the projection from the input data point to the nearest decision hyperplane. Our experiments based on $c$-Eval also suggest the existence of nearly-affine decision surfaces in several well-known image classifiers.
% 	\item We develop a low-complexity approach, namely $c$-Eval plot,  to evaluate and visualize explainer's quality. Experimental results show that $c$-Eval plot provides deep insights on explainers' performance under different sparsity constraints.
% 	\item We conduct extensive experiments on 8 explainers, using different image classifiers and perturbation schemes to validate the usage of $c$-Eval in evaluating explainers.
% \end{enumerate} 

{\bf Related Work.} Despite the recent development of interpretable machine learning, works focusing on evaluating explanations of local feature-based explainers are quite limited. To our knowledge, there are two independent works that can be considered to be directly relevant to $c$-Eval: the AOPC score ~\cite{Samek2017} and the log-odds score~\cite{Avanti2017}. %, and the Anchors explainer~\cite{Ribeiro2018AnchorsHM}. %In this section, we will provide a brief descriptions of these works and clarify the distinction between $c$-Eval and those works. %, which can be considered as explanations whose importance weights of all features are available. Specifically, 
 The AOPC score, which is introduced to evaluate heat-maps, is the average of the differences between the soft-outputs of the input image and those of some random perturbations. These random perturbations are generated sequentially based on the heat-maps on the input's features. Once may think to extend AOPC to evaluate explainers, such as mask-form explanation LIME; however, it is ambiguous due to an absence of the importance ordering. Furthermore, the AOPC needs a large number of random perturbations to make a stable random evaluation while computing $c$-Eval is a deterministic process requiring only one perturbation per evaluation. On the other hand, Shrikumar {\em et. al}~\cite{Avanti2017} use the log-odds score, measuring the difference between the input image and the modified image whose some pixels are erased, to evaluate explanations ~\cite{Avanti2017}. Given the weight importance on each pixel provided by the explanation, the pixels to be erased are chosen greedily from the one with the highest weight. The modified image is generated by continuing to erase more pixels until the predicted label of the modified image is different from that of the original image. However, the log-odds method is proposed without detailed analysis and it is only applicable to small gray-scale images like MNIST~\cite{lecun2010}.

{\bf Organization.} The rest of the paper is organized as follows. In Section~\ref{cEval}, we introduce the notations and formulates $c$-Eval, the unified metric to evaluate explanations of feature-based local explainers. Then, we describe how to compute $c$-Eval in Section~\ref{onnorm}. In Section~\ref{semantic}, we demonstrate the relationship between $c$-Eval and the importance of input features. Then, we propose the $c$-Eval plot, a visualization method based on $c$-Eval to examine explainers' behavior in Section~\ref{cplot}. Section~\ref{robustmodel} includes our demonstration of $c$-Eval on adversarial-robust models. %In Section ~\ref{related}, we discuss some recent works related to evaluating explanations. 
Our experimental evaluations on explanations to validate the usage of $c$-Eval are demonstrated in Section~\ref{simulations}. Finally, Section~\ref{conclusion} concludes the paper with a discussion on future directions.

\section{c-Eval of Explanation} \label{cEval}
In this section, we demonstrate the formulation of $c$-Eval metric in details. We consider a neural network as a function $f$ that accepts an input vector $\bm x \in \mathbb{R}^n$ and outputs a prediction $\bm y \in \mathbb{R}^m$. For a given vector $\bm x$, we use the notation $x_i$ to address the element $i^{\textup{th}}$ of vector $\bm x$. The label of a prediction $\bm y$ is denoted as $l = \arg \max_{1\leq j \leq m} y_j$. Given $g_f$, a feature-based local explainer on the classifier $f$, an explanation of prediction $f(\bm x)$ is a subset of features (elements) of $\bm x$. Formally, we write $e_{\bm x} = g_f(\bm x) \subseteq \bm x$. For convenience, we denote $e_{\bm x}$ the \textit{explanatory features} and $\bm x \setminus e_{\bm x}$ the \textit{non-explanatory features} of prediction $f(\bm x)$ made by $g_f$.

In feature-based explanations, explainer may simply return $e_{\bm x} = \bm x$ as an explanation for prediction $f(\bm x)$. We can interpret this answer as \textit{because the input is $\bm x$ so the prediction is $f(\bm x)$}. Even though this explanation is correct, it is not desirable since it neither gives us any additional information on the prediction nor strengthens our trust on the model's decision. A better answer is a smaller set of explanatory features that are important to the prediction. In fact, it is a common practice for explainers to impose cardinality constraints on $e_{\bm x}$ for more compact explanations~\cite{Marco2016, Ribeiro2018AnchorsHM}. Thus, when evaluating the quality of explanations, we assume that they are all subjected to the same cardinality constraint $| e_{\bm x}| \leq k$ for a fix integer  $k$.

We denote a perturbation scheme $h_{g_f}: \mathbb{R}^n \rightarrow \mathbb{R}^n $ of explanation $e_{\bm x}$ is a perturbation which only makes modification on features not in $e_{\bm x}$, i.e. non-explanatory features of $f(\bm x)$:
\begin{align}
h_{g_f}(\bm x)_i =x_i + \delta_i \quad \textup{where} \begin{cases}
\delta_i = 0 \quad \textup{if } x_i \in e_{\bm x}
\\ 
\delta_i \geq 0 \quad \textup{if } x_i \notin e_{\bm x}.
\end{cases} \label{hgf}
\end{align}
We call a perturbation $h_{g_f}$ of explanation $e_{\bm x}$ is a successful perturbation under the $p$-norm constraint $c$ if the label of prediction on $\bm x$ is changed after the perturbation and the $p$-norm distance between $\bm x$ and $h_{g_f}(\bm x)$ is not greater than $c$. Specifically, these conditions can be formulated as:
\begin{align}
&\arg \max_{1\leq j \leq m} f(h_{g_f}(\bm x)) \neq  l \nonumber \\
& || h_{g_f}(\bm x) - \bm x ||_p \leq c.  \label{ctolerance}
\end{align}
Based on this condition, we have the definition of $c$-Eval as follows:
\begin{definition}
	An explanation $e_{\bm x}$ of prediction $f(\bm x)$
     is $c$-Eval if no perturbing scheme $h_{g_f}$ of $e_{\bm x}$  that can change the model prediction on $\bm x$ while keeping the total distortion in $p$-norm less than or equal to $c$.
\end{definition}

Our intuition on $c$-Eval is that a good feature-based explanation is supposed to be $c$-Eval with high value of $c$. Because, if the features in $e_{\bm x}$ were important to the prediction $f(\bm x)$, the non-explanatory features should have minimal contribution to the prediction. As perturbation scheme $h_{g_f}$ is only allowed to perturb on non-explanatory features, $h_{g_f}$ must make significant changes to successfully alter the label of the prediction. Consequently, for a given explanation $e_{\bm x}$, the greatest value of $c$ in (\ref{ctolerance}) such that if there was no $h_{g_f}$ which successfully changes the prediction's label, it would imply the important of features in $e_{\bm x}$. Thus, we denote
\begin{align}
    & c_{f,\bm x} (e_{\bm x}) = \sup c \nonumber \\
     \textup{s.t.} \ & \nexists h_{g_f} \textup{ satisfying (\ref{ctolerance}) }. \label{cEvaldef}
\end{align}
In short, for every $c \leq c_{f,\bm x} (e_{\bm x})$, there is no perturbation scheme on non-explanatory features that can alter the label of prediction while keeping the total amount of distortion less than $c$. We call $c_{f,\bm x}(e_{\bm x})$ is the $c$-Eval of explanation $e_{\bm x}$. 

To this point, we have formulated the definition of $c$-Eval and described our intuition on the connection between $c$-Eval of an explanation and the importance of the explanatory features. Based on that connection, we propose to use $c$-Eval as a quantitative metric to evaluate the quality of explanations of neural networks. Before discussing on computing $c$-Eval in Section~\ref{onnorm} and strengthening the relationship between $c$-Eval and the important of explanatory features in Section~\ref{semantic}, we want to emphasize some properties of $c$-Eval and several remarks on the usage of $c$-Eval.

\begin{figure*}[h]
    \centering
  \includegraphics[width=0.9\textwidth]{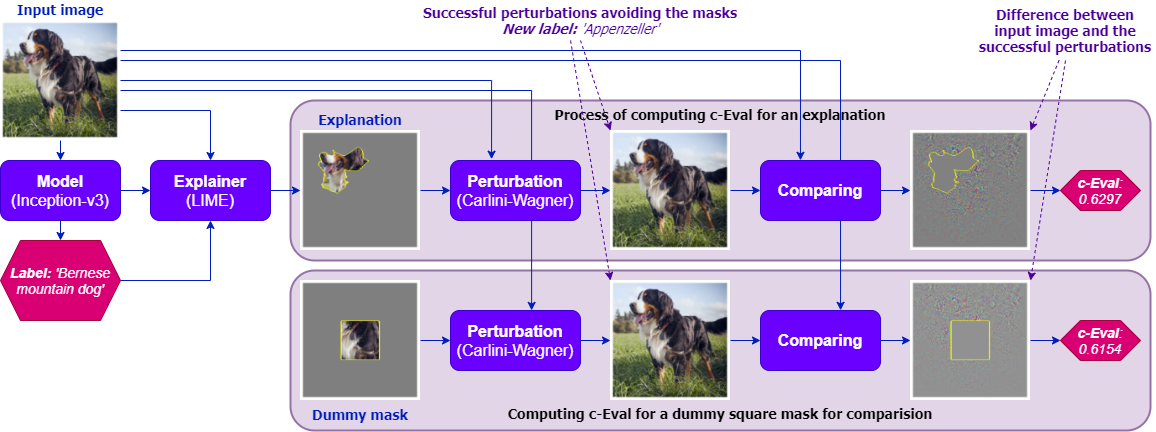}
  \caption{Example comparing the importance of features including in an explanation and features in a dummy mask using $c$-Eval.} \label{exampleprocess}
\end{figure*}

\textbf{Range of c-Eval.} When there is no element in the set of explanatory features, we have $c_{f,\bm x}(e_{\bm x}) = c_{f,\bm x}(\emptyset)$ is the minimum amount of perturbation onto all input's features to successfully change the original prediction. In this case, the successful perturbation $h_{g_f}$ will return a perturbation known as the \textit{minimally distorted adversarial examples} \cite{Carlini2017}. On the other hand, when explainer $g_f$ returns all features of the input image, there is no perturbation $h_{g_f}$ can alter the prediction's label and we set $c_{f,\bm x}(\bm x) = \infty$. 

\textbf{Size of explanations should be similar.} We limit the usage of $c$-Eval to explanations of the same or comparable sizes. The reason is an explainer can simply include a lot of unnecessary features in its explanation and trivially increase its $c$-Eval. However, this restriction does not prevent the usage of $c$-Eval in evaluating explanations of different explainers. In fact, we can always fix a compactness parameter $k$ (number of input features, number of pixels or number of image's segments as explanatory features) and take the top-$k$ important elements as an explanation. Therefore, when comparing different explainers using $c$-Eval, we will specify how compactness parameters of  explanations are chosen. For most experiments in this paper, $k$ is chosen to be $10\%$ of the number of input features.

\textbf{Normalize c-Eval among inputs.} Given a compactness parameter $k$, for different inputs $\bm x$, the amount of minimum distortion to make a successful perturbation can vary significantly. Hence, for meaningful statistical results, some experiments in this paper use the normalize ratio between $c$-Eval of $e_{\bm x}$ and $c$-Eval of empty explanation, i.e. $C_{f,\bm x}(e_{\bm x}) = c_{f,\bm x}(e_{\bm x})/ c_{f,\bm x}(\emptyset)$, to evaluate the quality of $e_{\bm x}$. 

\textbf{The choice of norm for $c$-Eval.} To our knowledge, there has been no research on which distance metric is optimal to measure the interpretability of explanations. There is also no consensus on the optimal distance metric of human perceptual similarity~\cite{Carlini2017}. Because of the followings reasons, we consider the $L_2$-norm, i.e. $p=2$, throughout this work: (i) $L_2$-norm has been used to generate explanation for neural networks' predictions~\cite{Marco2016}, (ii) our computation of $c$-Eval is related to the generation of adversarial samples, whose initial work~\cite{Christian2014} used $L_2$-norm, and (iii) there exists efficient algorithms to minimize $L_2$-norm in adversarial generation~\cite{foolbox2017, ding2019advertorch}. Even though we only study $c$-Eval in $L_2$-norm, the finding of a good distance metric is an important research question which we leave to the future works.

% Given an explanation, it is not straight-forward to compute its $c$-Eval by just using formula (\ref{cEvaldef}). In the next Section~\ref{onnorm}, we describe in details how $c$-Eval can be computed.

\section{Computing $c$-Eval} \label{onnorm}

Given an explanation, it is not straight-forward to compute its $c$-Eval by using formula (\ref{cEvaldef}). Instead, we solve for the successful perturbation scheme with the smallest distortion. Specifically, we compute $c$-Eval based on the following equivalent definition:
\begin{align}
    & c_{f,\bm x} (e_{\bm x}) = \inf c \nonumber \\
     \textup{s.t.} \ & \exists h_{g_f}. \textup{ satisfying (\ref{ctolerance}) }. \label{cEvaldef2}
\end{align}
Based on (\ref{cEvaldef2}), the $c$-Eval of explanation $e_{\bm x}$ can be obtained by solving for the minimum perturbation scheme $h_{g_f}$ on non-explanatory features. 

The computation processes of $c$-Eval can be summarized through an example shown in Fig.~\ref{exampleprocess}. Given an input image and an explanation for the prediction on that image, we compute the minimal distortion successful perturbation on non-explanatory features of that image using the "Perturbation" block. The $c$-Eval of the explanation is then the norm of the difference between the minimal distortion perturbation and the input image. In Fig.~\ref{exampleprocess}, we generate an explanation of LIME explainer for the prediction \textit{Bernese mountain dog} on the given input image. The explanation in this case includes roughly $10\%$ the total number of input pixels. After that, a perturbed instance $ h_{g_f}(\bm x)$ is generated using our modified version of Carlini-Wagner (CW) attack~\cite{Carlini2017} where the perturbation avoids the explanatory features. Then, the $c$-Eval is the norm of the difference between the input image and the perturbed instance. The reported $c$-Eval computed in the $L_2$-norm is $0.6297$. For the sake of demonstration, we construct a "dummy mask" of the same size as the LIME explanation, which include the center region of the original image. We consider this mask as an explanation for the prediction and compute the $c$-Eval for it, which is $0.6154$. Here, the $c$-Eval of LIME is larger than that of the "dummy mask", i.e. the amount of perturbation required to change the prediction while fixing the explanatory features of LIME is greater. This result is intuitive since we expect that LIME explanation should be better than a dummy square to explain the model's prediction.

% To this point, we want to mention the existence of recent developed Anchors explainer~\cite{Ribeiro2018AnchorsHM}, which seems to share similar viewpoint with $c$-Eval in term of important features. However, Anchor is a randomize explainer based on perturbing training data and its objective is clearly different from $c$-Eval's objective. 

For the computation of $c$-Eval, the only key step that requires a further specification is the "Perturbation" step (Fig.~\ref{exampleprocess}) to find a minimum perturbing scheme $h_{g_f}$ on non-explanatory features. We implement this step by modifying the CW attack so that it only performs perturbations on non-explanatory features. We select the CW attack for our implementation since it has been widely considered as the state-of-the-art algorithm generating minimal distortion adversarial samples of neural networks. In the CW attack, the algorithm solves for the optimal $\bm \delta \in [0,1]^n$ that minimizes the following objective:
\begin{align}
    &\mathcal{D}(\bm x, \bm x+ \bm \delta) + c. l(\bm x+ \bm \delta)
\end{align}
where $\mathcal{D}$ is a distance metric between the perturbation and the original image, $l$ is a loss function such that $l(\bm x+ \bm \delta) \leq 0$ if and only if the label of $\bm x + \bm \delta$ is different from the original label and $c > 0$ is a constant. Note that $\bm \delta$ also needs to satisfy the box-constraint $\bm x + \bm \delta \in [0,1]^n$ so that the perturbation is a valid input. Then, the optimal $\bm \delta$ is learnt via gradient-descents.

A simple modification of the CW attack so that it only conducts perturbations on non-explanatory features is blocking the backward steps on explanatory features in the gradient descents. However, the rate of convergence of a such modification may reduce significantly if many $\delta_i$ components with high gradients are blocked. We observe that the situation happens frequently as most existing explainers tend to include high-gradient components as explanatory features. 

To overcome this problem, we introduce perturbation variables $\bm \delta_{e_{\bm x}} \in [0,1]^{n - |e_{\bm x}|}$ representing perturbations on non-explanatory features. We then use a mapping $s: [0,1]^{n - |e_{\bm x}|} \rightarrow [0,1]^{n}$ that transforms the perturbation information in $\bm \delta_{e_{\bm x}}$ into $\bm \delta$. The mapping guarantees that for any explanatory feature $i$, $\delta_i = 0$. By using $s$, we can guarantee that the optimization steps focus on non-explanatory features. To solve for $\bm \delta_{e_{\bm x}}$, we use Adam~\cite{Adam2014} optimizer with the following objective:
\begin{align}
    \mathcal{D}(\bm x, \bm x+ s(\bm \delta_{e_{\bm x}})) + c. l(\bm x+ s(\bm \delta_{e_{\bm x}})).
\end{align}

One drawback of CW attack is the high running-time complexity. However, from the perspective of $c$-Eval, the minimal distortion perturbation might not necessary to evaluate explanations. For example, consider that we have an algorithm to find a successful perturbation on non-explanatory features of $\bm x$. If $e_{\bm x}$ is important to the prediction, it will be difficult for the algorithm to find a successful perturbation by perturbing only on $\bm x \setminus e_{\bm x}$. The intuition here is very similar to the definition of $c$-Eval in previous section. The only difference is in the space of the perturbation schemes. Thus, we extend our definition of $c$-Eval to the ``$c$-Eval with respect to a class of perturbing scheme $\mathcal{H}$'' as follows.
\begin{definition}\label{onH}
    An explanation $e_{\bm x}$ of prediction $f(\bm x)$
     is $c$-Eval with respect to the class of perturbing schemes $\mathcal{H}$ if there is no perturbing scheme $h_{g_f} \in \mathcal{H}$ of $e_{\bm x}$ that can change the model prediction on $\bm x$ while keeping the total distortion in $p$-norm less than or equal to $c$.
\end{definition}

Definition~\ref{onH} helps us avoid the difficulty in finding the minimum-distortion perturbation scheme $h_{g_f}$. Instead of examining all perturbations scheme satisfying the $p$-norm constraint within distance $c$, we can focus on the optimal $h_{g_f}$ in a much smaller set of perturbation schemes $\mathcal{H}$. By narrowing down the choices of $h_{g_f}$, we make the computation of $c$-Eval tractable without much loss in performance. Specifically, we propose to focus on the set of perturbations generated by the Gradient-Sign-Attack (GSA)~\cite{Goodfellow2015}, and the Iterative-Gradient-Attack (IGA)~\cite{Alexey2016} due to their low running time complexity. Given an image $\bm x$, GSA sets the perturbation $\bm x'$ as
\begin{align}
    \bm x' = \bm x - \epsilon.\textup{sign}(\nabla J_l(\bm  x ) ),
\end{align}
where $J_l$ is the $l$ component of the loss function used to train the neural network and $\epsilon$ is a small constant. On the other hand, IGA initializes $\bm x'^{(0)} = \bm x$ and update it iteratively as
\begin{align}
    \bm x'^{(i+1)} = \textup{clip}_{\bm x,\epsilon} \left( \bm x'^{(i)} - \alpha.\textup{sign}(\nabla J_l(\bm  x'^{(i)} ) ) \right)
\end{align}
where the clip function ensures that $\bm x'^{(i)}$ is in the $\epsilon$-neighborhood of the original image. To adopt GSA and IGA into the context of $c$-Eval where the perturbation is on non-explanatory features, we simply block the backward step of gradient-descent algorithm on explanatory features.

\begin{figure}[ht]
	\centering
% 	\vspace{-4mm}
    \begin{subfigure}{.15\textwidth}
		\centering
		\includegraphics[scale=0.25]{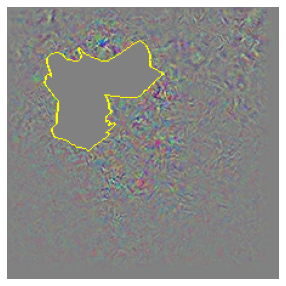}
		\caption{LIME-CW}
	\end{subfigure}
	\begin{subfigure}{.15\textwidth}
		\centering
		\includegraphics[scale=0.25]{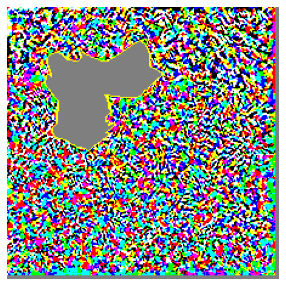}
		\caption{LIME-GSA}
	\end{subfigure}
	\begin{subfigure}{.15\textwidth}
		\centering
		\includegraphics[scale=0.25]{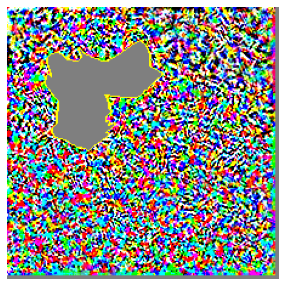}
		\caption{LIME-IGA}
	\end{subfigure}
	\begin{subfigure}{.15\textwidth}
		\centering
		\includegraphics[scale=0.25]{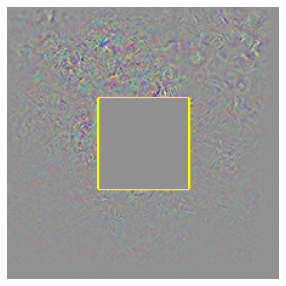}
		\caption{Dummy-CW}
	\end{subfigure}
	\begin{subfigure}{.15\textwidth}
		\centering
		\includegraphics[scale=0.25]{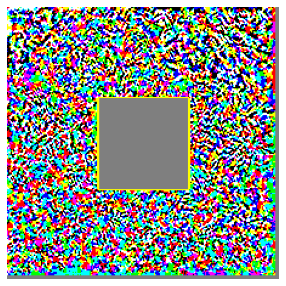}
		\caption{Dummy-GSA}
	\end{subfigure}
		\begin{subfigure}{.15\textwidth}
		\centering
		\includegraphics[scale=0.25]{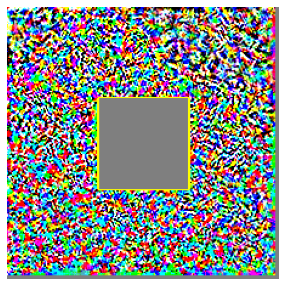}
		\caption{Dummy-IGA}
	\end{subfigure}
	\caption{Distortions between perturbations and the original images. The notation 'LIME' and 'Dummy' stand for LIME explanation and dummy explanation in experiment of Fig.~\ref{exampleprocess}.}
	\label{distortion}
% 	\vspace{-4mm}
\end{figure}

Fig.~\ref{distortion} shows the distortions between successful perturbations generated by different attacks. Here, the experiment setup including the model and the input image are the same as in the experiment of Fig.~\ref{exampleprocess}. The $L_2$-norm of the distortions generated by GSA and IGA on LIME explanation are $1.3120$ and $0.9804$, respectively. The corresponding $c$-Eval for the dummy mask are $1.2962$ and $0.9696$. We can see that the distortions in GSA and IGA are more spreading out due to the nature of the attacks, which constitutes higher total distortions. Even though the distortions in GSA and IGA are larger than those computed by CW attack, their results still imply that LIME explanation is better than the dummy mask and align with our intuition on the explanation's quality.

\begin{figure}[h]
		\centering
		\includegraphics[scale=0.45]{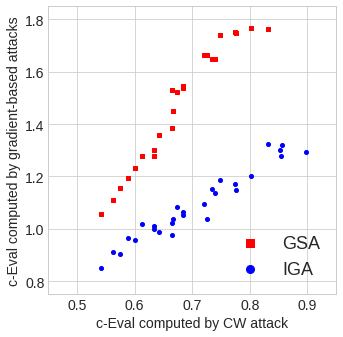}
		\caption{Scatter plot of $c$-Eval computed by gradient-based attacks vs $c$-Eval computed by CW attack.}
		\label{gradandcw}
\end{figure}

In Fig.~\ref{gradandcw}, we provide 
the scatter plot of $c$-Eval of 30 explanations in Inception-v3 computed by different perturbations methods. From the plot, we are able to see the strong correlations of $c$-Eval computed by GSA as well as IGA to $c$-Eval obtained from CW attack. Based on these correlations, we will use GSA and IGA instead of CW attack to compute $c$-Eval for some experiments in this work due to their low running time complexity.

\section{c-Eval and the importance of features} \label{semantic} 
This section illustrates a relationship between $c$-Eval and the importance of features returned by local explainers. We first demonstrate this relationship in multi-class affine classifiers. We show that $c$-Eval determines the minimum distance from the explained data point (the input image) to the nearest decision hyperplane in a lower-dimension space restricted by the choice of explanatory features. A high $c$-Eval implies that the chosen explanatory features are more aligned with the minimum projection's direction, i.e. they are key features determining the prediction on the data point. We further extend the analysis of $c$-Eval to general non-affine classifiers. Our experiments based on $c$-Eval suggest an existence of nearly-affine decision surfaces in several well-known classifiers.

% An explanation with high $c$-Eval then must contain important features since those features align with the minimum projection's direction.

% Specifically, when we perturb those features, the data point is moved along the projection's direction. When the perturbation is large enough, the data point is moved into the other half of the classifier's decision hyperplane and its prediction is changed. In the following, using the definition of $c$-Eval, we will show that the metric determines the minimum distance from the examined data point to the nearest decision hyperplane in a lower-dimension space, whose dimensions are restricted by the choice of the explanation. Consequently, an explanation with high $c$-Eval is expected to contain important features since those features must be aligned with the minimum projection's direction.  \mt{This is quite lenthy introduction of a section. Details should be in IV.A.}

\subsection{c-Eval in affine classifiers.} 

% {\color{blue}In affine classifier, the prediction space are divided by multiple decision hyperplanes. Given an input data point and its prediction made by the affine classifier,  

% Hence, the important features of the classifier's prediction on a given data point are features whose dimensions are more aligned with the projection from the data point to the nearest decision hyperplane. If those features are kept unchanged, to alter the classifier's prediction, the data point must be moved along a different direction which results in a significantly larger total distortion. In $L_2$-norm, this distortion is the length of the projection from the data point to the nearest decision hyperplane in the "non-important" features space. As we will show $c$-Eval of an explanation is that length in the non-explanatory features space, an explanation with high $c$-Eval must contain important features since keeping those features unchanged gives us high total distortion.}

We consider an affine classifier $f(\bm x) = \bm W^T \bm x + \bm b$ where $\bm W$ and $\bm b$ are given model's parameters. Given an explanation $e_{\bm x}$, $c$-Eval is the solution of the following program:
\begin{align}
	\min& \ ||\bm \delta||_2 \label{distanceOpt}\\
	\textup{s.t}& \  \exists j: \bm w^T_j(\bm x + \bm \delta) + b_j \geq \bm w^T_{j_0}(\bm x + \bm \delta) + b_{j_0}, \nonumber\\
	& \ \forall i \in e_{\bm x}, \delta_i = 0, \nonumber
\end{align}
where $\bm w_j$ is the $j^{\textup{th}}$ column of $\bm W$, $j_0=  \arg \max_{ j} f(\bm x)$ is the original prediction and $\bm \delta$ is the vector of $\delta_i$ defined in (\ref{hgf}). 

\begin{figure}[ht]
	\centering
%	\vspace{-6mm}
	\hspace*{-7mm}
	\includegraphics[scale=0.25]{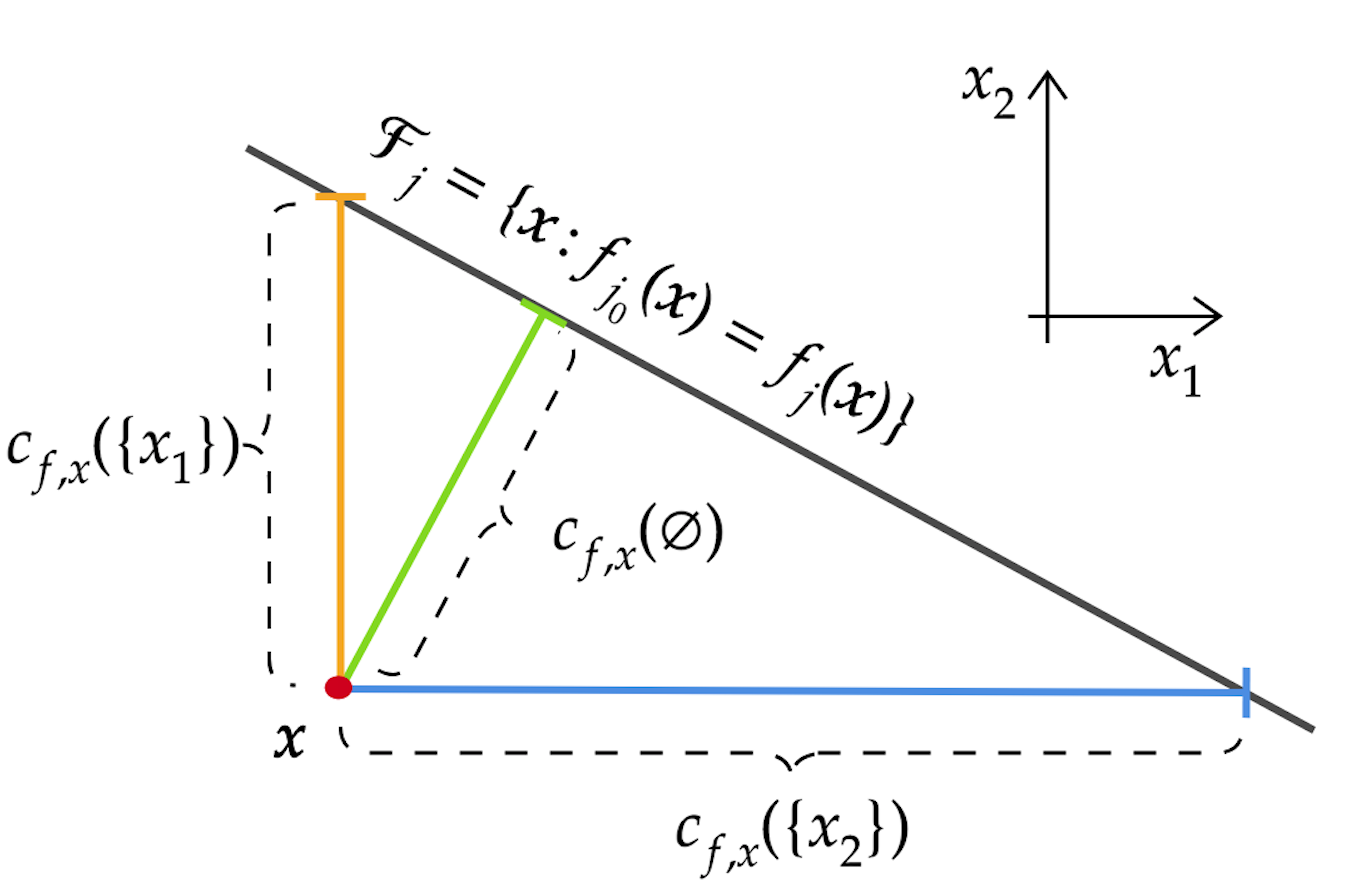}
	\caption{$c$-Eval in 2D affine classifier. Explanation $\{x_2\}$ is better than $\{x_1\}$ since the distance from $\bm x$ to hyperplane $\mathcal{F}_j$ without changing $x_2$ is larger than that distance without changing $x_1$.}
	\label{affine}
\end{figure}

When $e_{\bm x} = \emptyset$, there is no restriction on entries of $\bm \delta$. The optimization program (\ref{distanceOpt}) computes the distance between $\bm x$ and the complement of convex polyhedron $P$:
\begin{align}
	P =  \bigcap_{j=1}^m \{ \bm x: f_{j_0}(\bm x) \geq f_j(\bm x)\}, \label{poly}
\end{align}
where $\bm x$ is located inside $P$. The optimal $c_{f,\bm x}(\emptyset)$ of (\ref{distanceOpt}) is a distance from $\bm x$ to the closest decision hyperplane $\mathcal{F}_j = \{\bm x: f_{j_0}(\bm x) = f_j(\bm x) \}$ of $P$. For the sake of demonstration, Fig.~\ref{affine} describes an example in 2-dimension space where $c_{f,\bm x}(\emptyset)$ is plotted in the green line.

For each explanatory feature in $e_{\bm x}$, the optimization space of (\ref{distanceOpt}) is reduced by one dimension. The optimization program (\ref{distanceOpt}) then solves for the shortest distance from $\bm x$ to the complement of polyhedron $P$ in lower dimension. In 2-dimension space as depicted in Fig.~\ref{affine}, under an assumption that $\mathcal{F}_j$ is also the closest hyperplane of $P$ to $\bm x$, $c_{f,\bm x}(e_{\bm x} = \{x_1\})$ is the distance from $\bm x$ to $\mathcal{F}_j$ when the feature $x_1$ is unchanged. Similarly, the $c$-Eval $c_{f,\bm x}(e_{\bm x} = \{x_2\})$ is the length of the blue line in the figure. In this case, allowing changing $x_2$ is easier to alter the original prediction $j_0$ than $x_1$, i.e. $c_{f,\bm x}(\{x_1\}) < c_{f,\bm x}(\{x_2\})$. It implies that $x_2$ is more important to the prediction than $x_1$. 

To this point, we see that under the affine assumption on classier $f$, $c$-Eval of an explanation is the length of the projection from the data point to the decision hyperplanes in the space of non-explanatory features. Therefore, the explanation with high $c$-Eval contains features whose dimensions are more aligned with the shortest distance vector from the data point to the decision hyperplane. Thus, $c$-Eval reflects the importance of features in the explanation.

%It is also important to point out another interesting property of $c$-Eval in affine classifier. Given an explanation $e_{\bm x}$, we have shown that $c_{f,\bm x}(e_{\bm x})$ is the distance from $\bm x$ to $\mathcal{F}_j$ without changing features in $e_{\bm x}$. Thus,  $c_{f,\bm x}(\bm x \setminus e_{\bm x})$ is the distance from $\bm x$ to $\mathcal{F}_j$ without changing the complement of $e_{\bm x}$.

\subsection{c-Eval in general non-affine classifiers.} 
For general classifiers, the set $P$ in equation (\ref{poly}) describing the region of prediction $j_0$ is no longer a polyhedron. However, our observation based on $c$-Eval suggests that many well-known image classifiers might be nearly affine in a wide-range of local predictions. Therefore, it is still applicable to  evaluate models' explanations using $c$-Eval. 
\begin{figure}[ht]
    \vspace{-5mm}
	\centering
	\includegraphics[scale=0.22]{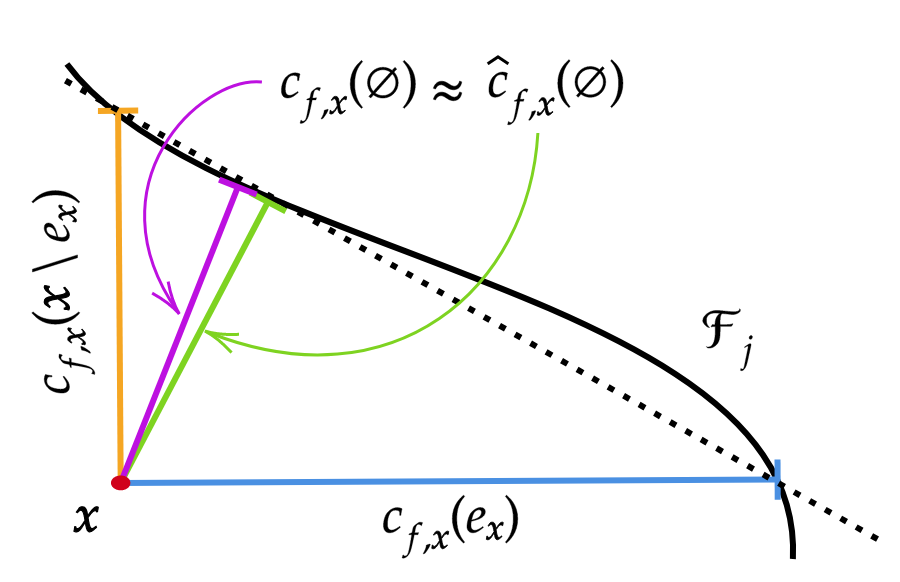}
	\caption{$c$-Eval in non-linear classifier. When $f$ is nearly affine, $ c_{f,\bm x}(\emptyset) \approx \hat{c}_{f,\bm x}(\emptyset)$.}
	\label{nonaffine}
\end{figure}

Our observation is based on a property of $c$-Eval in affine classifier. Given an explanation $e_{\bm x}$, we have shown that $c_{f,\bm x}(e_{\bm x})$ is the distance from $\bm x$ to $\mathcal{F}_j$ without changing features in $e_{\bm x}$. Similarly,  $c_{f,\bm x}(\bm x \setminus e_{\bm x})$ is the distance from $\bm x$ to $\mathcal{F}_j$ without changing the complement of $e_{\bm x}$. As in the case of 2-dimension in Fig~\ref{affine}, $c_{f,\bm x}(\emptyset)$ is the height to the hypotenuse of the right triangle whose sides are  $c_{f,\bm x}(e_{\bm x})$ and  $c_{f,\bm x}(\bm x \setminus e_{\bm x})$. Thus, we have the following equalities:
\begin{align}
&\frac{1}{c_{f,\bm x}(\emptyset)^2} = 	\frac{1}{c_{f,\bm x}(e_{\bm x})^2} + 	\frac{1}{c_{f,\bm x}(\bm x \setminus e_{\bm x})^2}\\
\leftrightarrow \ & c_{f,\bm x}(\emptyset) = \frac{1}{\sqrt{	1/{c_{f,\bm x}(e_{\bm x})^2} + 	1/{c_{f,\bm x}(\bm x \setminus e_{\bm x})^2}}}, \label{approxcempty}
\end{align}
for any explanation $e_{\bm x}$. We denote the expression on the right-hand-side of (\ref{approxcempty}) by $\hat{c}_{f,\bm x}(\emptyset) $. 

\begin{figure}[ht]
	\centering
	\vspace{-5mm}
	\hspace*{-30mm}
	\begin{subfigure}{.15\textwidth}
		\centering
		\includegraphics[scale=0.3]{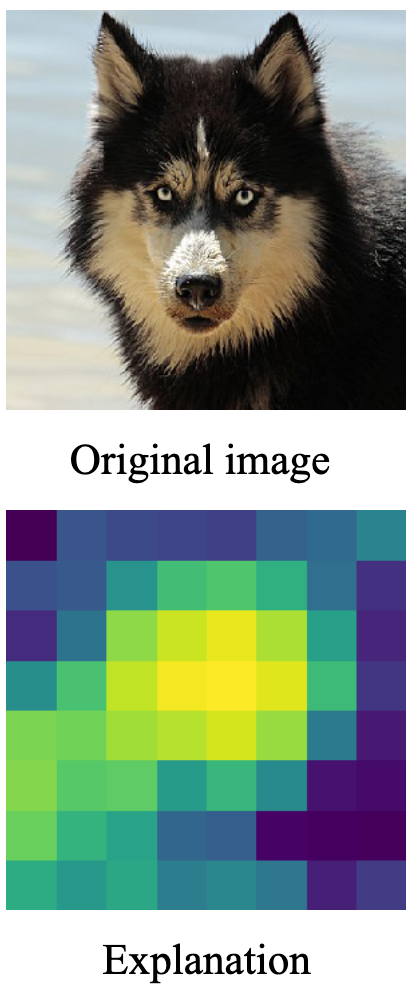}
	\end{subfigure}
	\begin{subfigure}{.15\textwidth}
		\centering
		\includegraphics[scale=0.24]{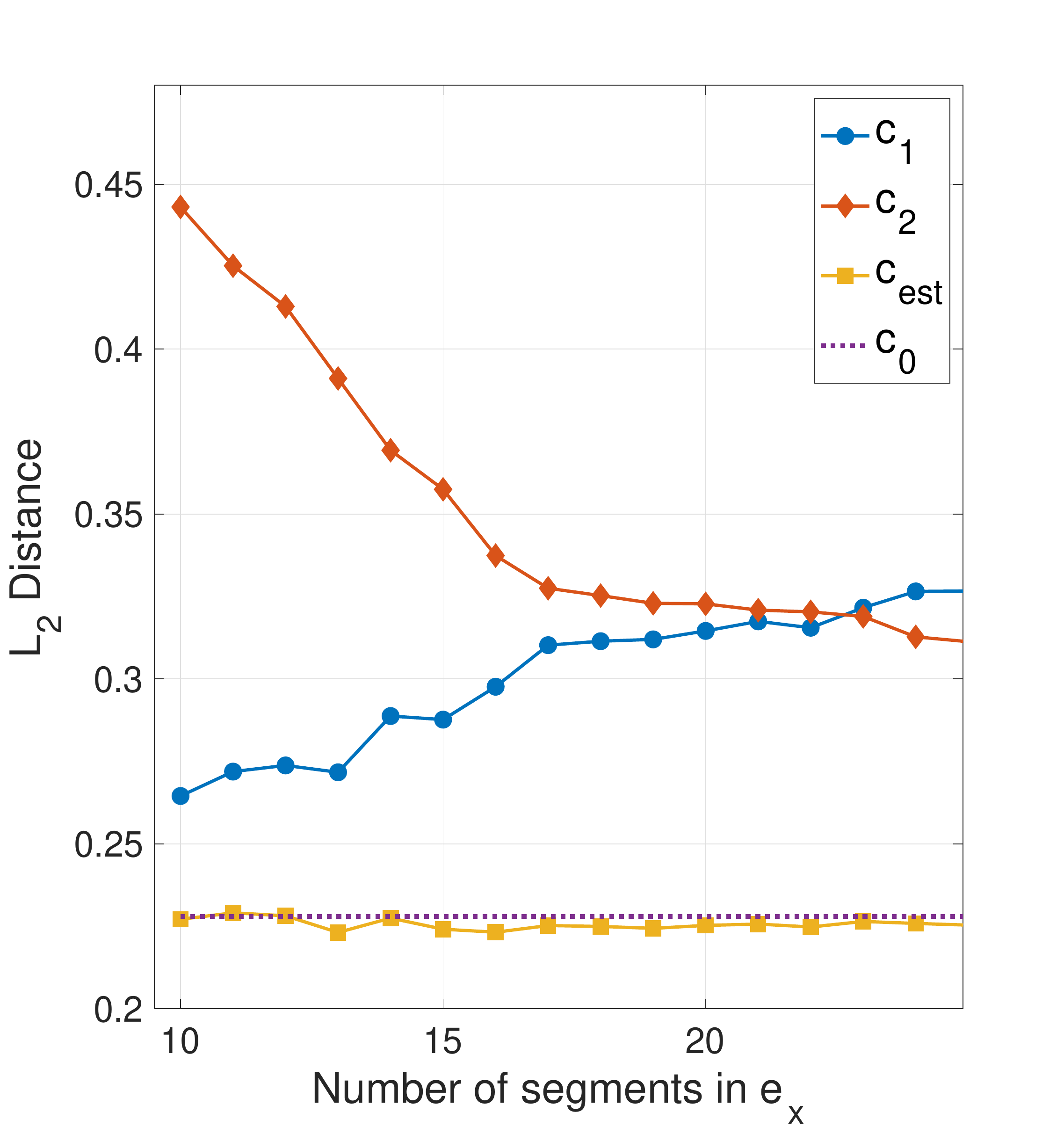}
	\end{subfigure}
	\caption{Example of nearly affine instance on Inception-v3. Here, $c_1,c_2,c_{\textup{est}}$ and $c_0$ are $c_{f,\bm x}(e_{\bm x}),c_{f,\bm x}(\bm x \setminus e_{\bm x}), \hat{c}_{f,\bm x}(\emptyset)$ and $ c_{f,\bm x}(\emptyset)$ respectively. Since $ c_{f,\bm x}(\emptyset) \approx \hat{c}_{f,\bm x}(\emptyset)$ for all number of segments from the explanation, we might infer that the decision surface is nearly affine in this example.}
	\label{nearaffine}
\end{figure}

For non-linear classifiers $f$, equation (\ref{approxcempty}) does not hold in general. However, if the decision surface $\mathcal{F}_j$ is nearly affine, we should have $ c_{f,\bm x}(\emptyset) \approx \hat{c}_{f,\bm x}(\emptyset)$ for all $e_{\bm x}$ as described in Fig.~\ref{nonaffine}. By testing different classifiers, we observe that this necessary condition hold for many data points of common image classifiers such as Inception-v3~\cite{Inception2015}, VGG19~\cite{Simonyan2014} and ResNet50~\cite{He2015}. For example, we generate a $8\times 8$ GCam explanation on the Inception-v3 and iteratively compute $c_{f,\bm x}(e_{\bm x})$ and  $c_{f,\bm x}(\bm x \setminus e_{\bm x})$ using CW attack. Here, we vary the number of explanatory features $k$ in $e_{\bm x}$ and compute the corresponding $\hat{c}_{f,\bm x}(\emptyset)$ using equation (\ref{approxcempty}). The results are plotted in Fig.~\ref{nearaffine}. The value of $c_{f,\bm x}(\emptyset)$ is drawn using the purple straight-dot-line for reference. We can see that the two lines for $c_{f,\bm x}(\emptyset) $ and $\hat{c}_{f,\bm x}(\emptyset)$ are close to each other.

For VGG19 and ResNet50, we use the same input image as for Inception-v3 with GSA to reduce the running time complexity. Note that the $L_2$ distance is computed based on the input space of each model. We can see that $\hat{c}_{f,\bm x}(\emptyset)$ and $ c_{f,\bm x}(\emptyset)$ are close to each others in both cases. It is interesting that different models share this same property, which encourage us to use $c$-Eval to evaluate explanations of those classifiers.
\begin{figure}[ht]
	\centering
	\vspace{-4mm}
	\begin{subfigure}{.24\textwidth}
		\centering
		\includegraphics[scale=0.19]{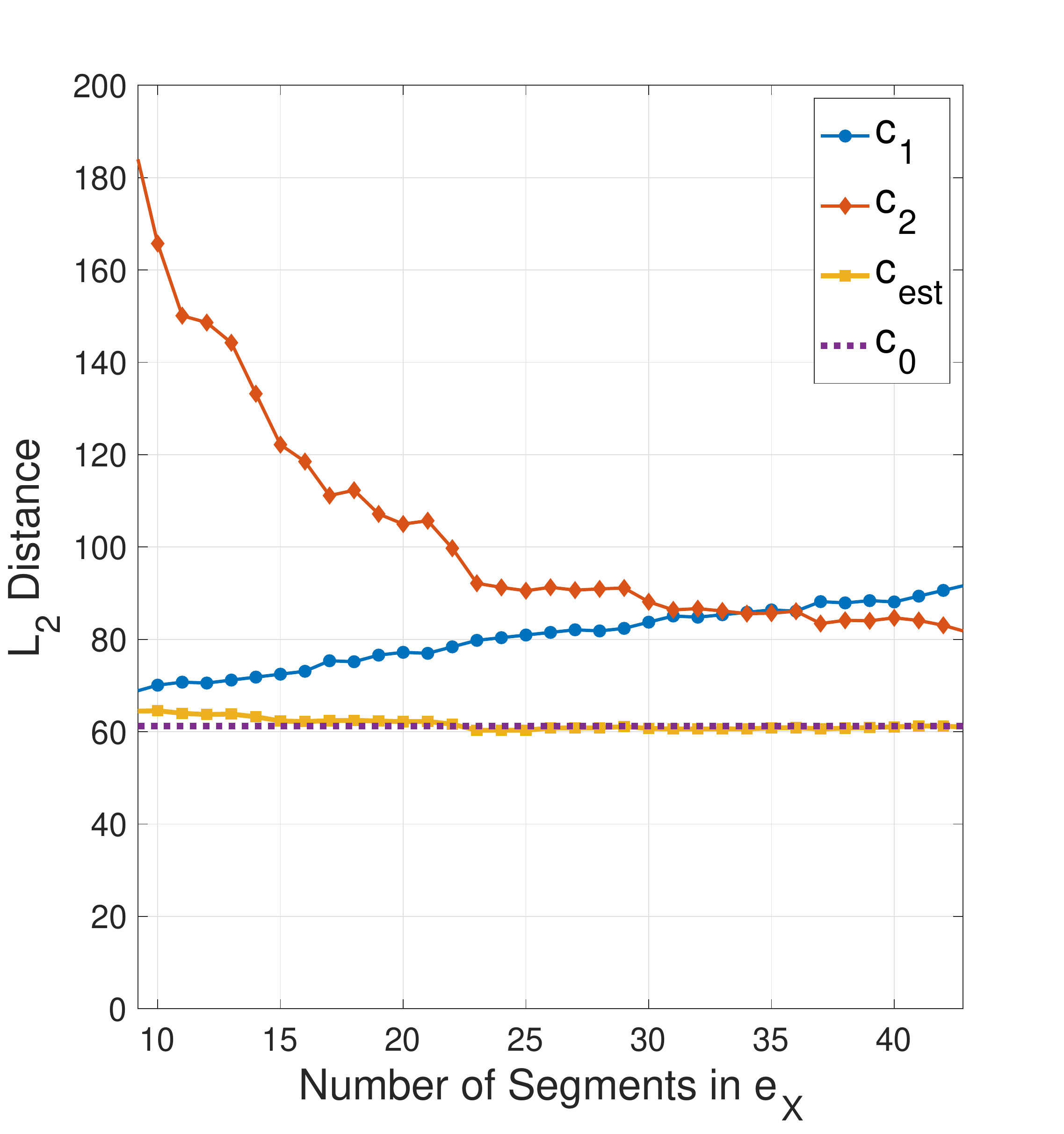}
		\caption{VGG19}
	\end{subfigure}
	\begin{subfigure}{.24\textwidth}
		\centering
		\includegraphics[scale=0.19]{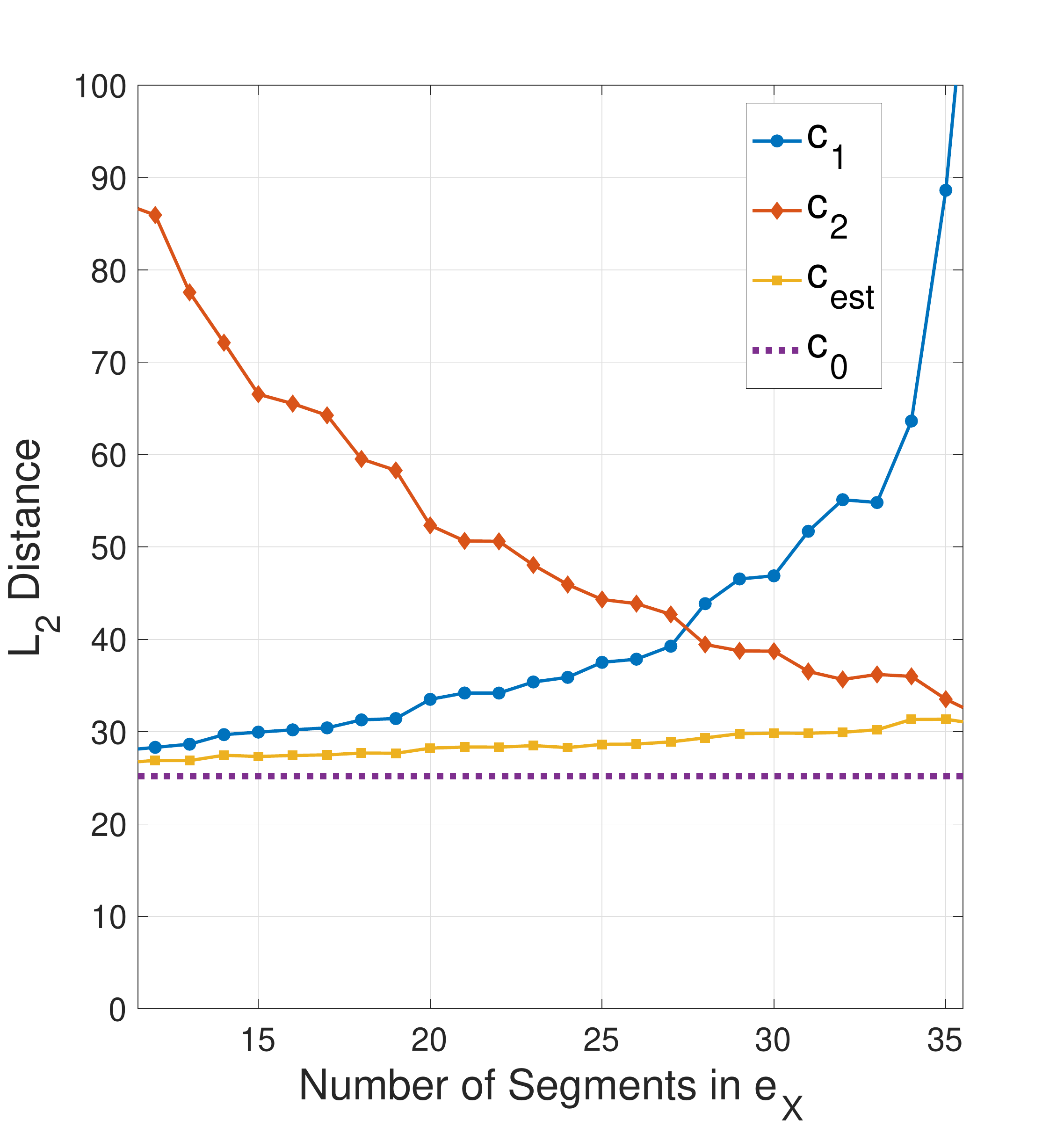}
		\caption{ResNet50}
	\end{subfigure}
	\caption{The condition $c_{f,\bm x}(\emptyset) \approx \hat{c}_{f,\bm x}(\emptyset)$ also holds for VGG19 and ResNet50, which also suggests the existence of nearly-affine decision surfaces in these models.}
	\label{resnetandvgg}
	\vspace{-4mm}
\end{figure}

\section{Beyond c-Eval} \label{beyond}
%To this point, the concept of $c$-Eval is introduced to evaluate explanations. 
In Subsection~\ref{cplot}, we introduce the $c$-Eval plot, which is a visualization of explainers' behavior on a given input based on $c$-Eval. Using examples on LIME explainer, we demonstrate that $c$-Eval plot not only help us determine the appropriate tuning parameters for LIME but also strengthen the usage of $c$-Eval in evaluating the importance of explanatory features. Since $c$-Eval relies on generating successful perturbations, Subsection~\ref{robustmodel} discusses $c$-Eval's behavior in adversarial-robust models. We show that $c$-Eval computed in adversarial-robust models are strongly correlated with its counterpart in non-robust models, which implies that $c$-Eval can be adopted in adversarial-robust models.

%it raises concerns on $c$-Eval's behavior in adversarial-robust models - models that are trained to be more resistant to perturbations~\cite{Goodfellow2015}. In Subsection~\ref{robustmodel}, we heuristically show that the $c$-Eval computed in adversarial-robust models are strongly correlated with its counterpart in non-robust models, which implies that $c$-Eval can be adopted in adversarial-robust models.

\subsection{c-Eval plot} \label{cplot}
In Section~\ref{cEval}, we restrict the $c$-Eval analysis on explanations of similar sizes. That restriction is just for fair comparison among explanations of different explainers. Given an explainer, by varying the number of explanatory features $k$, we obtain a sequence of explanations with their corresponding $c$-Eval. Therefore, on a given input image, each explainer will be associated with a sequence of $c$-Eval values. By plotting this sequence as a function of $k$, we can observe the behaviors of explainers on that input and evaluate their validity. We call the resulting plot the $c$-Eval plot.

\begin{figure}[ht]
    % \vspace{-5mm}
	\centering
	\includegraphics[scale=0.26]{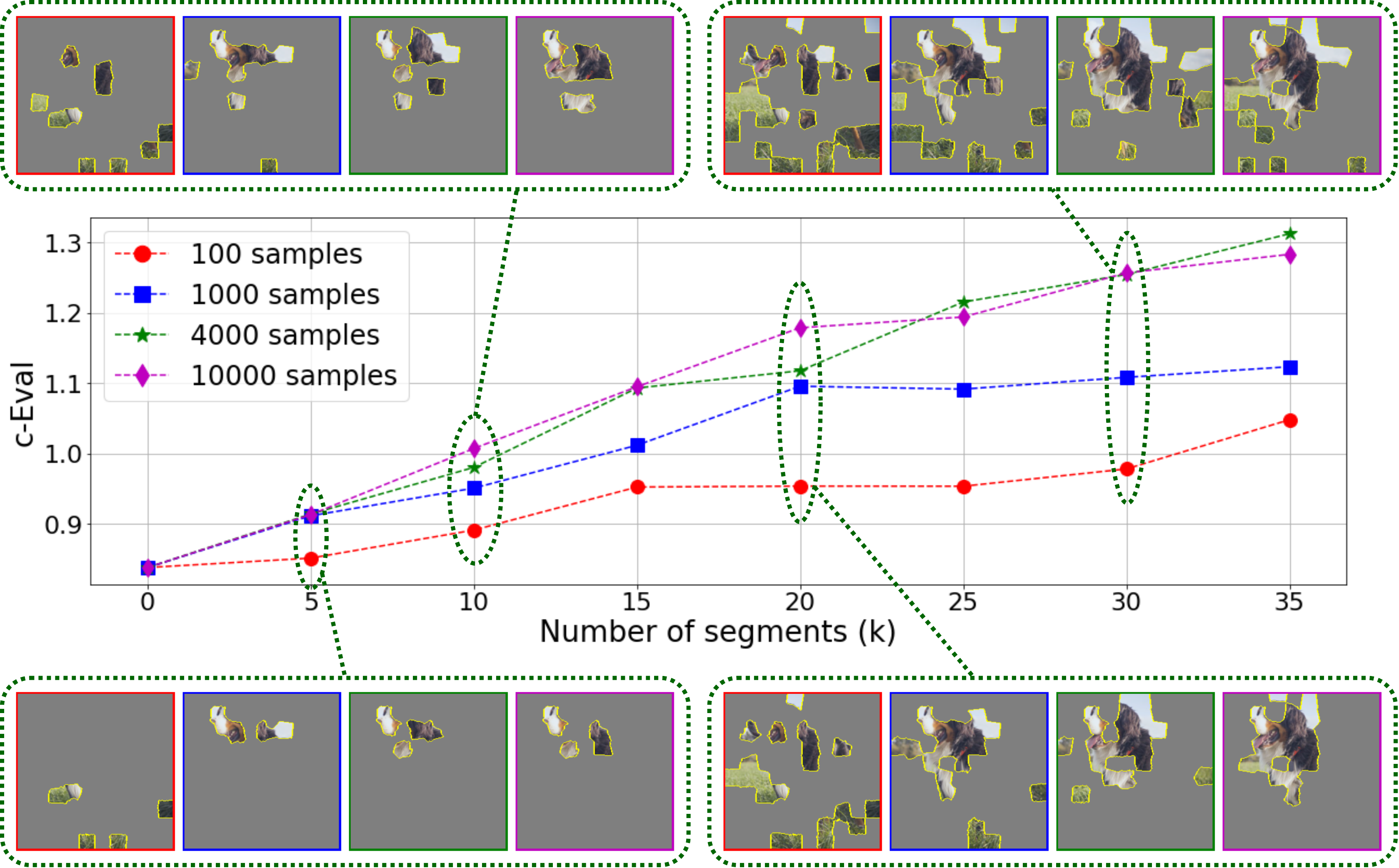}
		\caption{$c$-Eval plot of LIME with different sample rates. Higher sample rates result in better explanations and $c$-Eval plot reflects that expectation.}
	\label{cvs}
\end{figure}

In the followings, we provide an example of $c$-Eval plot and heuristically demonstrate that the $c$-Eval plot correctly evaluates explanations and helps us understand the behavior of the examined explainer on a given input. Here, we consider the LIME explainer with different number of samplings. In LIME, the sampling size determines how many perturbations are conducted in finding the explanation. The higher the number, the better the explanation and the higher the running time complexity~\cite{Marco2016}. Since there is no concrete rule on how this parameter should be chosen, how can we verify that a LIME explanation is free from under-sampling error? On the other hand, if the number of samplings is reduced, is the explanation still faithful to the prediction? We show how $c$-Eval plot can help us address these problems.

The experiments are conducted on Inception-v3 with the input image as in the experiment in Fig.~\ref{exampleprocess}. We first segmentized the input image into $100$ feature segments. Then, we explain them using LIME explainer with $100$, $1000$, $4000$ and $10000$ samples. We plot the the sequences $\{c_{f, \bm x}(e^{k}_{\bm x})\}_{k=1}^n$, i.e. the $c$-Eval plot in fig.~\ref{cvs}. For references, we also provide the explanations with number of segments $k=5,10,20$ and $30$ for each setting of LIME (red for $100$, blue for $1000$, green for $4000$ and purple for $10000$ samples). The figure shows a distinct gap in $c$-Eval among explanations of LIME. As can be seen,  %This result first supports the usage of $c$-Eval plot to evaluate the explainers' quality: 
the higher the number of samples, the higher the explanations quality and also the higher the $c$-Eval. Additionally, using $c$-Eval plot, we can deduce that there is not much improvement in the explanations' quality by increasing the number of samples from $4000$ to $10000$. This example shows that $c$-Eval can be used as a metric to support automatically tuning of explainer's parameters. It also helps us gain trust in LIME in the sense that, if we aim for top-$5$ important features among $100$ features, LIME with $2000$ samples might be reliable since there is not much gain in $c$-Eval by increasing that number from $1000$ to $10000$. 

\subsection{c-Eval on adversarial-trained models} \label{robustmodel}

Since $c$-Eval is computed based on adversarial generation, there might be several concerns regarding the applications of $c$-Eval on adversarial-robust models. First, as adversarial-robust models are more resistant to perturbations, is it viable to generate successful perturbations on robust models? Second, if we are able to obtain those perturbations, are the $c$-Eval of the corresponding explanations reliable? Here, we address those concerns through experiments on MNIST dataset using LeNet model~\cite{lecun1998}. Specifically, we show that the $c$-Eval on non-robust and robust models have strong correlation. This correlation implies that the behaviors of $c$-Eval are similar on non-robust and robust models.

We use Advertorch \cite{ding2019advertorch}, a Python toolbox for adversarial robustness research, to train three LeNet classifiers on MNIST dataset. The first model, denoted as non-robust model, is trained normally on the dataset. The second model is alternatively trained between images from MNIST and the corresponding adversarial samples generated at each iteration. Here, the normalized $L_2$-norm distortion between each adversarial sample and its original image is bounded by $\epsilon = 0.3$. The third model is trained in the same manner as the second model where the bound $\epsilon$ is set to $0.5$. All three classifiers archive more than $95\%$ accuracy on test set. For the two adversarial-trained models, their accuracy on adversarial samples are all greater than $94\%$. 

Using  4000 images in the test set, we generate their predictions made by the three LeNet classifiers and the corresponding top-$10\%$ LIME explanations. For all three classifiers, we are able to obtain the successful perturbations using IGA and the corresponding $c$-Eval for all explanations. The successful perturbations for all inputs of adversarial-trained models can be computed because the models are only robust against adversarial with bounded distortion. In $c$-Eval, the perturbations are not limited by the amount of distortion. %at all. 

\begin{figure}[h]
		\centering
		\includegraphics[scale=0.45]{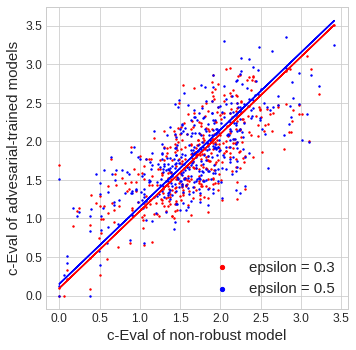}
		\caption{Correlation between $c$-Eval on non-robust and adversarial-trained models.}
		\label{scatter}
\end{figure}
	
Fig.~\ref{scatter} is the scatter plot of $c$-Eval of the three classifiers. We only plot $300$ data points for the sake of demonstration. The computed Pearson correlations between $c$-Eval of the first model and those of the other two adversarial-trained models are $0.765$ and $0.764$ respectively. We asset this is a fairly high correlation when we take into account that these are three separate models. In fact, on average, less than $75\%$ of the explanatory features are shared between non-robust model and any of the two robust ones. We also observe the model with higher bound in the distortion in the training has higher average $c$-Eval. This aligns with our intuition on how $c$-Eval is computed.

% We adopt the Keras classifier on CIFAR10 and train the correspond robust classifier by the Adversarial Robustness Toolbox~\cite{art2018}. The training data of the original classifier is 5000 images while the training data for robust model is 10000 images - 5000 from the training data and 5000 from their corresponding adversarial instances generated by CW-Attack. We compute the classifiers' predictions on 100 images of CIFAR10, whose prediction's labels on both classifiers are identical. We then generate the corresponding LIME explanations of the predictions on the non-robust classifier. The $c$-Eval of those explanations on both classifiers are computed by CW-Attack and plotted in Fig.~\ref{robust}. Note that we use the explanations of the non-robust classifier for the robust one since we want to alleviate the affect of explainers on this experiment. The examined explanations contain 10$\%$ or 30$\%$ features of the original images respectively. 

\section{Experimental results}\label{simulations}

In this section, we use $c$-Eval to experimentally evaluate explanations generated by different feature-based local explainers on small gray-scale hand-writing images MNIST~\cite{lecun2010} and large color object images Caltech101~\cite{FeiFei2004}. We also provide experimental results showing that evaluations of explanations based on $c$-Eval on MNIST dataset align with previous results obtained from log-odd scoring function~\cite{Avanti2017}, which is specifically designed for MNIST dataset only. To demonstrate the statistic behavior of $c$-Eval on large number of samples, the reported $c$-Eval is not precisely $c_{f,\bm x}(e_{\bm x})$ but the ratio of $c_{f,\bm x}(e_{\bm x})$ over $c_{f,\bm x}(\emptyset)$. This ratio is indicated by the notation C$(e_x)/$C$_0$ in the legend of each figure. The ground-truth quality rankings of explanations are obtained from previous results in assessing explainer's performance using human-based experiments~\cite{Scott2017,Avanti2017}. The studied classifier models and explainers are selected based on those previous experiments accordingly. The system specifications and the codes for our implementations are specified in subsection~\ref{sys}. For reference, we also provide experiments on small-size color image dataset CIFAR10~\cite{Krizhevsky2009}, which can be found in Appendix~\ref{CIFAR10app}.

\subsection{System specifications and source code} \label{sys}
Our experiments in this paper are conducted in Python. The computing platform is a Linux server equipped with two Intel Xeon E5-2697 processors supporting 72 threads. Our system memory comprises twelve 32 GB DDR4 sticks, each operates at 2400 MHz. 
% The source code for our experiments can be found at \cite{codeIML}.

\subsection{Simulations on MNIST dataset}\label{mnist}

For the MNIST dataset~\cite{lecun2010}, we study 8 different feature-based local explainers: LIME~\cite{Marco2016}, SHAP~\cite{Scott2017}, GCam~\cite{Ramprasaath2016}, DeepLIFT (DEEP)~\cite{Avanti2017}, Integrated Gradients~\cite{Sundararajan2016}, Layerwise Relevance Propagation (LRP)~\cite{Bach2015}, Guided-Backpropagation (GB)~\cite{Springenberg2014} and Simonyan-Gradient (Grad)~\cite{Simonyan2013}. Followings are brief descriptions of these explainers.

In LIME, the importance of each picture segment is approximated with a heuristic linear function using random perturbation. SHAP, which relies on the theoretical analysis of Shapley value in game theory, assigns each pixel a score indicating the importance of that pixel to the classifier's output. Since SHAP is a generalized version of LIME, we expect SHAP explanation to be more consistent with the classifier than LIME, hence SHAP's $c$-Evals are expected to be higher statistically. Previous work~\cite{Scott2017} also provided human-based experiments to support this claim. DeepLIFT, Integrated Gradients, LRP, GB and Grad are backward-propagation methods to evaluate the importance of each input neuron to the final output neurons of the examined classifier. Previous experiment results using log-odds function in~\cite{Avanti2017} suggest that  GB and Grad perform worse than the other three in MNIST dataset. The final studied explainer GCam is an image explainer designed specifically for fully-connected convolutional networks. It exploits the last convolution layer to explain the model's prediction. Since GCam is not designed for classifiers of low-resolution images, we expect its performance and the corresponding $c$-Eval in the MNIST dataset are limited.

\begin{figure}[ht]
% 		\vspace*{-3.0mm}
% 		\hspace*{-3.0mm}%
		\centering
		\begin{subfigure}{.24\textwidth}
			\centering
			\includegraphics[scale=0.2]{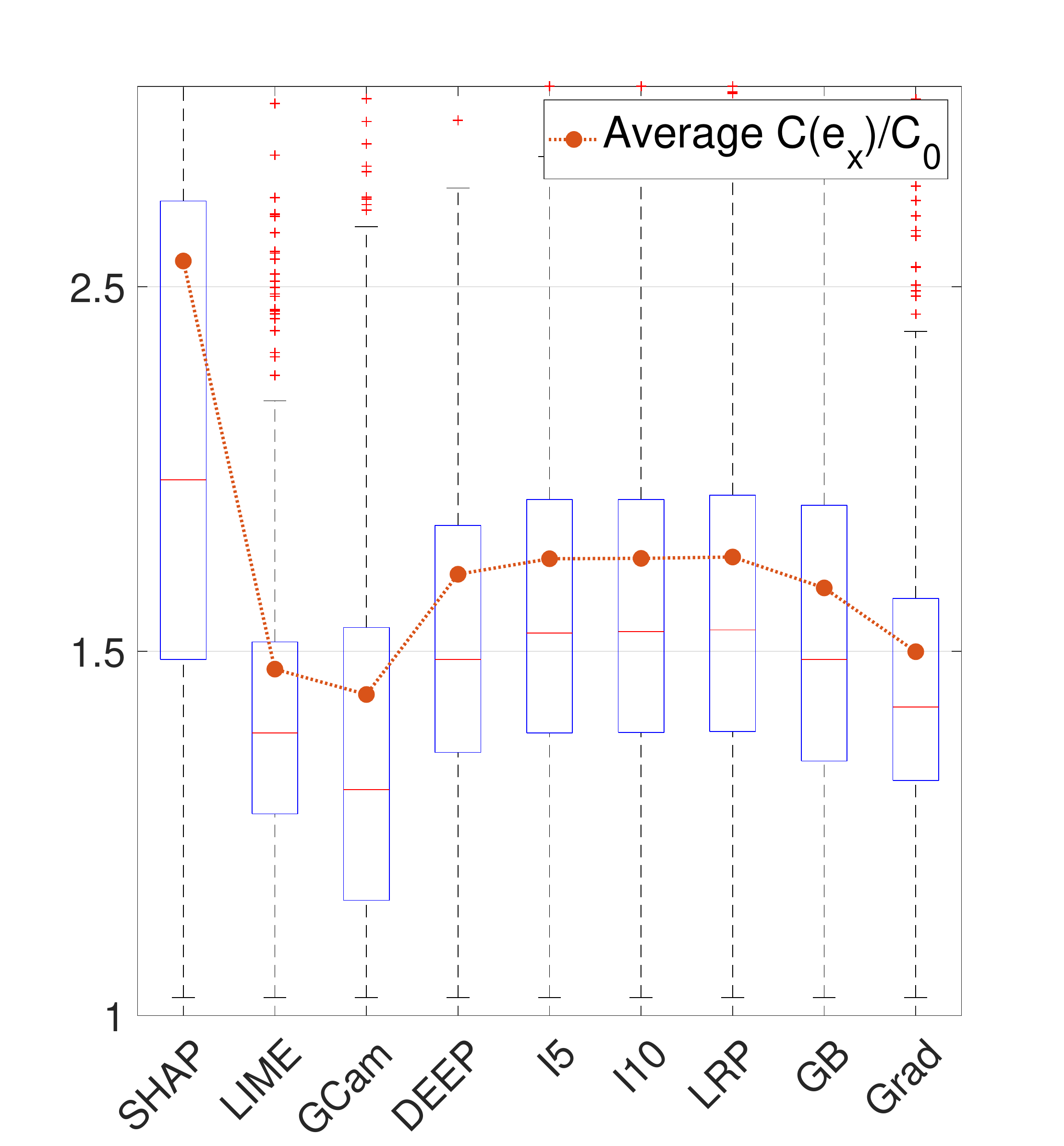}
			\caption{GSA on classifier 1.}
			\label{MNIST1000SH}
		\end{subfigure}
		\begin{subfigure}{.24\textwidth}
			\centering
			\includegraphics[scale=0.2]{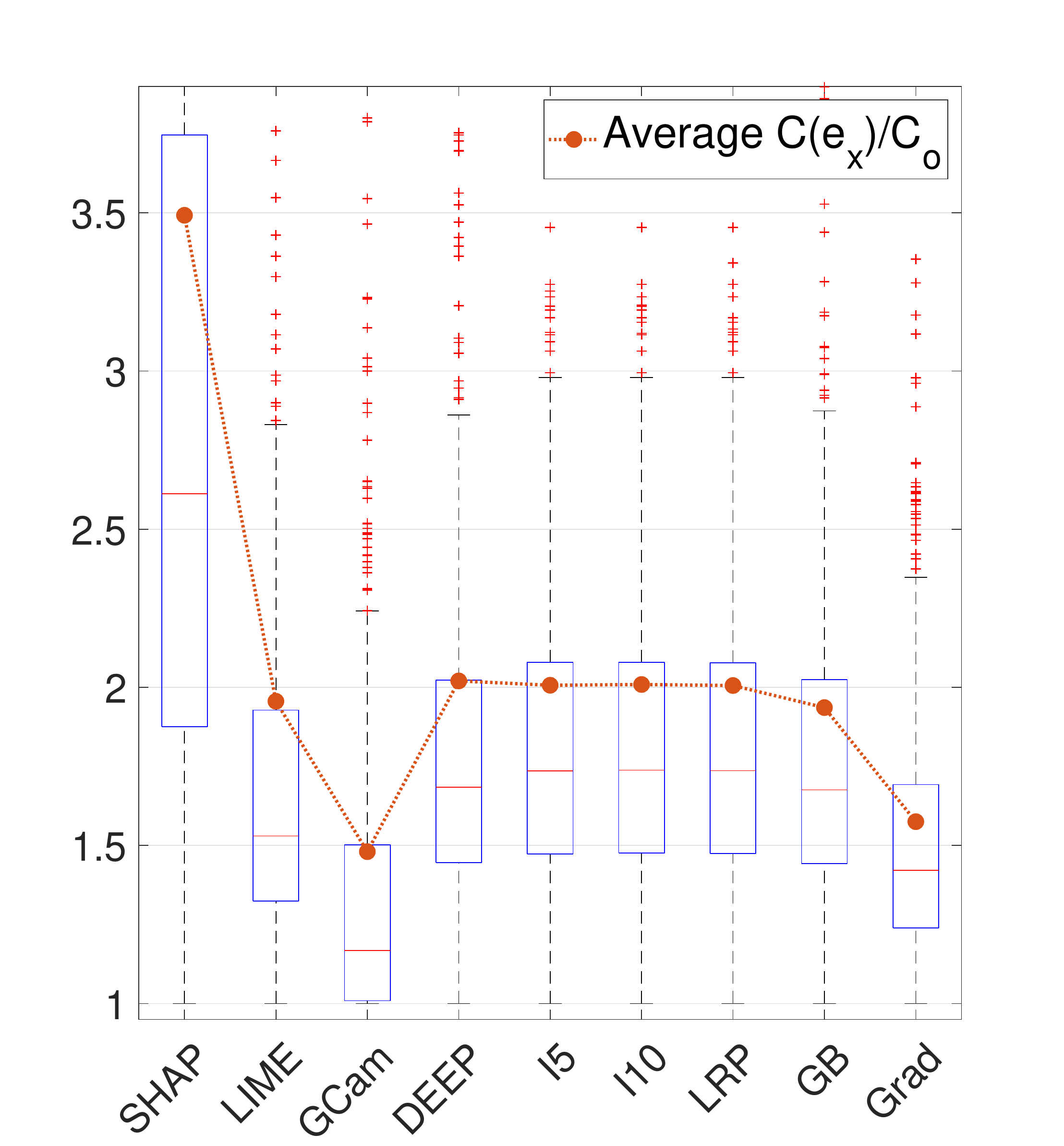}
			\caption{GSA on classifier 2.}
			\label{MNIST1000DL}
		\end{subfigure}
		\begin{subfigure}{.24\textwidth}
			\centering
			\includegraphics[scale=0.2]{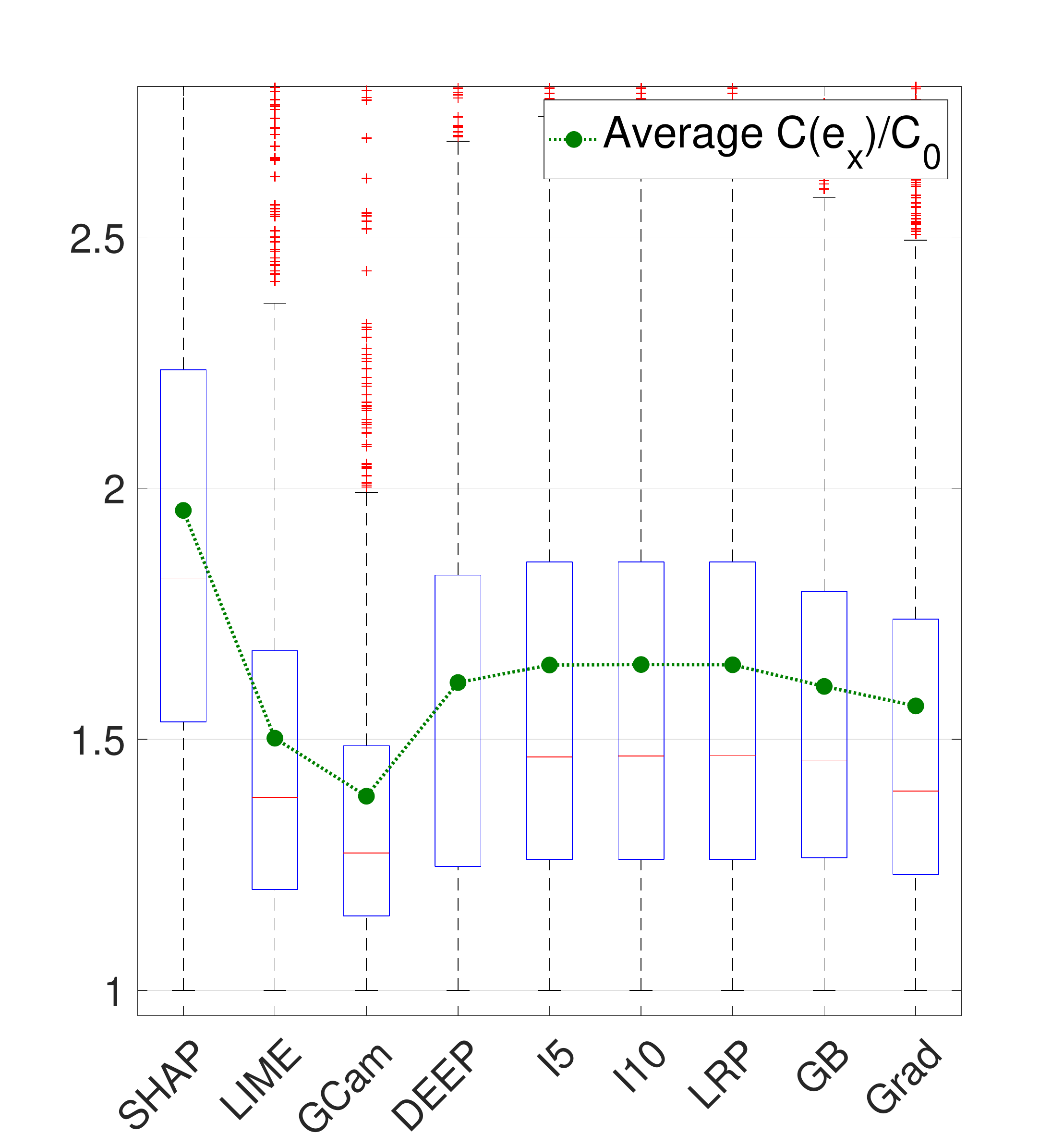}
			\caption{IGA on classifier 1.}
			\label{MNIST1000SHDiff}
		\end{subfigure}
		\begin{subfigure}{.24\textwidth}
			\centering
			\includegraphics[scale=0.2]{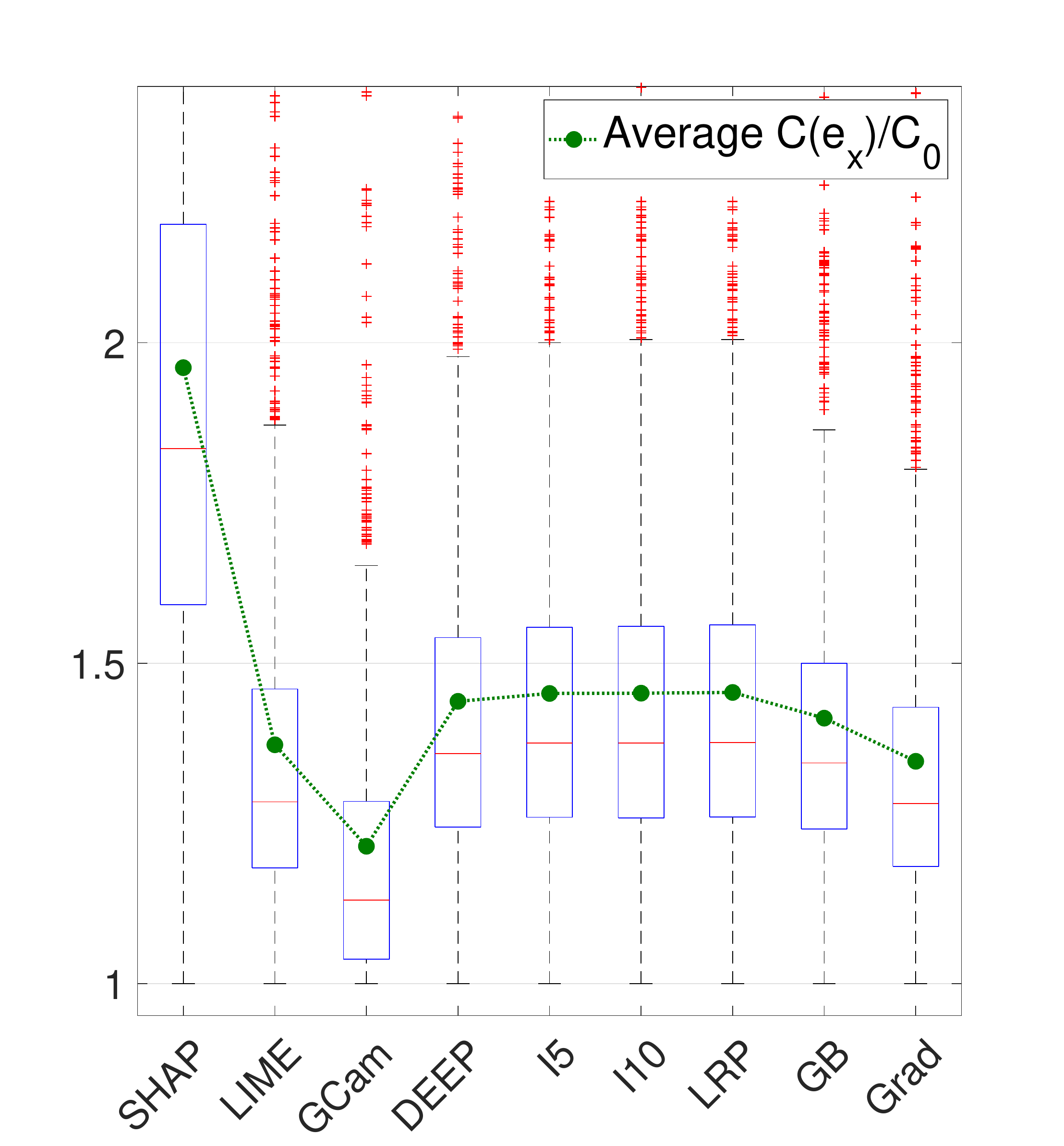}
			\caption{IGA on classifier 2.}
			\label{MNIST1000DDiff}
		\end{subfigure}
		\caption{We conduct the experiments for $8$ explainers on $1000$ images of MNIST dataset. Figure shows the distributions and the averages of $c$-Eval for $8$ explainers on classifier 1 provided by~\cite{Scott2017} and on classifier 2 provided by~\cite{Avanti2017}.}
		\label{dist}
% 		\vspace*{-5.0mm}
	\end{figure}

% \begin{figure}[ht]
% 	\centering
% 	\vspace{-2mm}
% % 	\hspace*{-8mm}
% 	\begin{subfigure}{.49\textwidth}
% 		\centering
% 		\includegraphics[scale=0.225]{Distribution_500_MNIST_9ex_0317_Smodel.eps}
% 		\caption{GSA classifier 1.}
% 		\label{MNIST1000SH}
% 	\end{subfigure}
% 	\begin{subfigure}{.49\textwidth}
% 		\centering
% 		\includegraphics[scale=0.225]{Distribution_500_MNIST_9ex_0317_Dmodel.eps}
% 		\caption{GSA classifier 2.}
% 		\label{MNIST1000DL}
% 	\end{subfigure}
% % 	\hspace*{-8mm}
% 	\begin{subfigure}{.49\textwidth}
% 		\centering
% 		\includegraphics[scale=0.225]{Distribution_1000_MNIST_9ex_0317_Smodel_Iattack.eps}
% 		\caption{IGA classifier 1.}
% 		\label{MNIST1000SHDiff}
% 	\end{subfigure}
% 	\begin{subfigure}{.49\textwidth}
% 		\centering
% 		\includegraphics[scale=0.225]{Distribution_1000_MNIST_9ex_0317_Smodel_Iattack.eps}
% 		\caption{IGA classifier 2.}
% 		\label{MNIST1000DDiff}
% 	\end{subfigure}
% 	\caption{Experiments for $8$ explainers on $1000$ images of MNIST dataset. Figure shows the distributions and the averages of $c$-Eval for $8$ explainers on two different classifiers.}
% 	\label{dist}
% 	\vspace{-5mm}
% \end{figure}

Our experiments on the MNIST dataset are conducted in pixel-wise manner, i.e. the outputs of explainers are image pixels. For each input image, each explainer except LIME is set to return $10\%$ the number of image pixels as explanation. For LIME, since the algorithm always returns image segments as explanations, we set the returned pixels to be as close to $10\%$ of the total number of pixels as possible. On another note, the implementations of LRP are simplified into Gradient$\times$Input based on the discussion in \cite{Avanti2017}. Some example explanations of images from MNIST are plotted in Fig. \ref{examples_MNIST}, shown in Appendix~\ref{exMC}. The $c$-Eval and the statistical results of explanations are reported in Fig.~\ref{dist}.  %With each explanation, we compute the corresponding $c$-Eval and report the statistical results in Fig.~\ref{dist}. %Followings are the details of the obtained results in Fig.~\ref{dist}.

\textit{Experiments on different classifiers:} Figs.~\ref{MNIST1000SH} and~\ref{MNIST1000DL} are the distributions of $c$-Evals of $1000$ images in MNIST dataset on classifier 1 provided by~\cite{Scott2017} and classifier 2 provided by~\cite{Avanti2017}. The notation I5 and I10 indicate the Integrated-Gradient method with 5 and 10 interpolations~\cite{Sundararajan2016}. We can see that the evaluation based on $c$-Eval is consistent between classifiers as well as previous attempts of evaluating explainers in~\cite{Scott2017} and~\cite{Avanti2017}. For the consistency in the behavior of $c$-Eval and log-odds function in~\cite{Avanti2017}, please see the discussion in subsection~\ref{sublogodds}.

\textit{Experiments on different gradient-based perturbation schemes:} Figs.~\ref{MNIST1000SHDiff} and~\ref{MNIST1000DDiff} demonstrate the usage of IGA instead of GSA as in experiments of Figs.~\ref{MNIST1000SH} and~\ref{MNIST1000DL}. Comparing the distributions in Fig.~\ref{MNIST1000SHDiff} to Fig.~\ref{MNIST1000SH} and Fig.~\ref{MNIST1000DDiff} to Fig.~\ref{MNIST1000DL}, we observe the relative $c$-Eval of explainers are similar on both perturbation schemes and consistent with our experiment in Fig.~\ref{gradandcw}. Thus, the computed $c$-Evals using IGA also reflect the explainers' performance. Finding optimal perturbation schemes resulting in a good measurement of $c$-Eval is not considered in this work; however, the experiments suggest that we can use non-optimal perturbation scheme to obtain reasonable measurement of $c$-Eval. 

\subsection{Simulations on Caltech101 dataset}

For experiments on large images, we study the performance of LIME, SHAP, GCam, DeepLIFT on 700 images in Caltech101 dataset~\cite{FeiFei2004} with the VGG19 classifier~\cite{Simonyan2014}. As LIME, SHAP, and GCam explainers are designed for medium-size to large-size images, we expect they should outperform DeepLIFT. Furthermore, the results from~\cite{Scott2017} implies SHAP should perform better than LIME. On the other hand, as GCam are designed for fully-connected convolution networks (e.g. VGG19), we expect its relative performance in these experiments to be much better than in previous experiments on MNIST dataset.

\begin{figure}
	\centering
	\begin{subfigure}{.225\textwidth}
		\centering
		\includegraphics[scale=0.21]{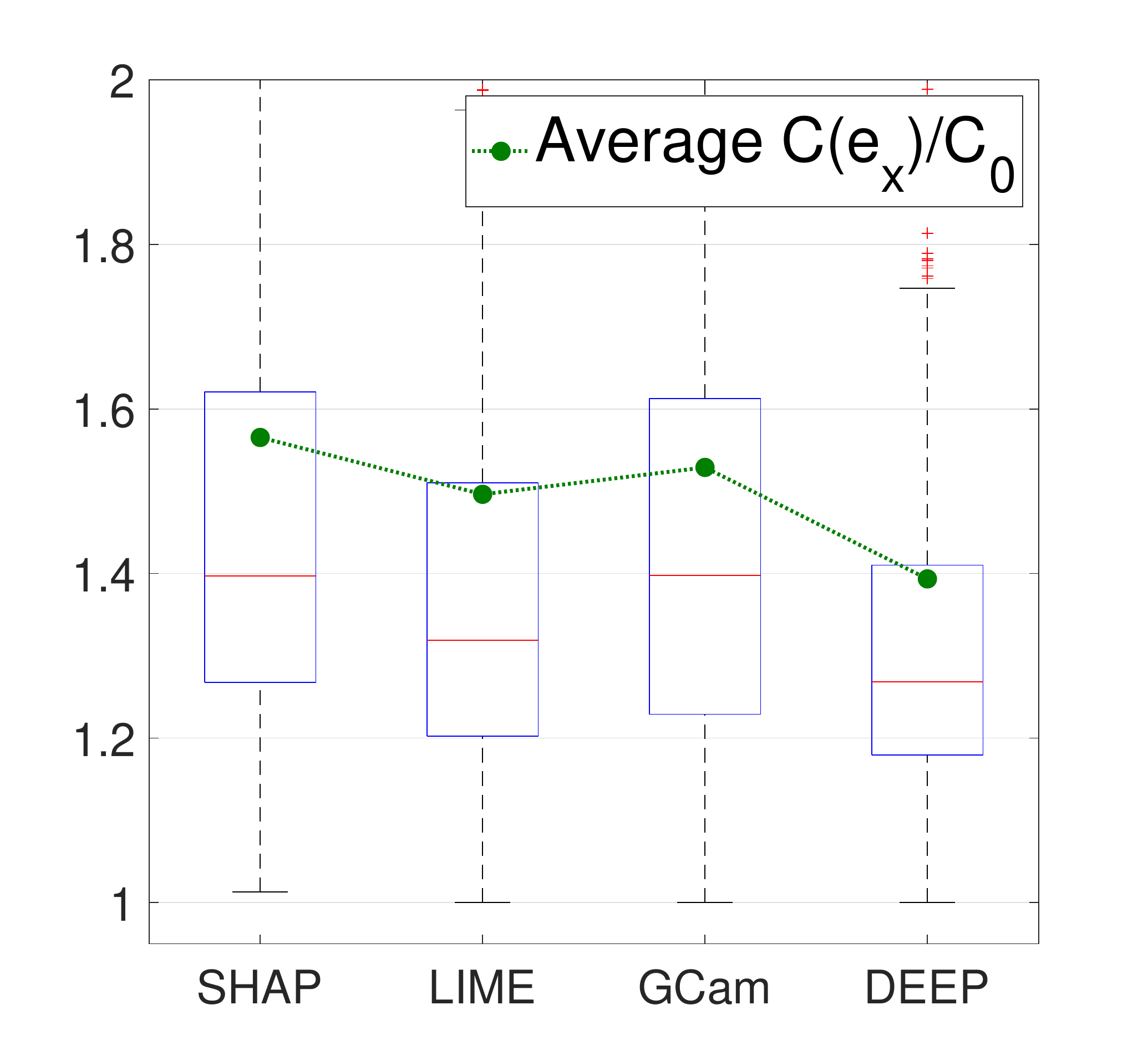}
		\caption{GSA.}
		\label{Caltech101vgg}
	\end{subfigure}
	\begin{subfigure}{.225\textwidth}
		\centering
		\includegraphics[scale=0.21]{Distribution_700_Caltech_Iterative_0317-eps-converted-to.pdf}
		\caption{IGA.}
		\label{Caltech101vggI}
	\end{subfigure}
	\caption{Distributions of $c$-Eval computed by GSA and IGA for four explainers in Caltech101 dataset.}
	\label{Caltech101}
	\vspace{-2mm}
\end{figure}

In these experiments, we use segment-wise features. Since the returned features of many explainers are importance weights of pixels, we need to convert them into a subset of image segments as explanations for fair comparison. We first segment each image into segments and then sum up the importance weights of all pixels inside each segment. We finally select the top $k$ segments with maximum sum-weight as the segment-wise explanation of the studied explainer. For the results in Fig.~\ref{Caltech101} each explainer returns the explanation with the number of segments roughly covers about $20\%$ of the original input image. Some examples of explanations in these experiments are shown in Fig.~\ref{Caltechapp} of Appendix~\ref{exMC}. 

The computed $c$-Eval in our experiments on Caltech101 are reported in Fig.~\ref{Caltech101vgg} and Fig.~\ref{Caltech101vggI}. Here, we use GSA and IGA to compute $c$-Eval respectively. We can observe that the statistical behavior of $c$-Eval aligns with our expectation on the performance of all four explanation method on this dataset. For the improvement of GCam and the degradation of DeepLIFT from the MNIST dataset and CIFAR10 to the Caltech101 dataset, we suggest readers to focus on the differences in quality of explainers among Fig.~\ref{examples_MNIST}, Figs.~\ref{examples_CIFAR} and Figs.~\ref{Caltechapp} in Appendix~\ref{exMC}.

\subsection{Similarity of c-Eval and log-odds functions in MNIST} \label{sublogodds}
	
		\begin{figure}[h]
		%\vspace*{-3.0mm}
		%\hspace*{-2.0mm}%
		\centering
		\begin{subfigure}{.235\textwidth}
			\centering
			\includegraphics[scale=0.19]{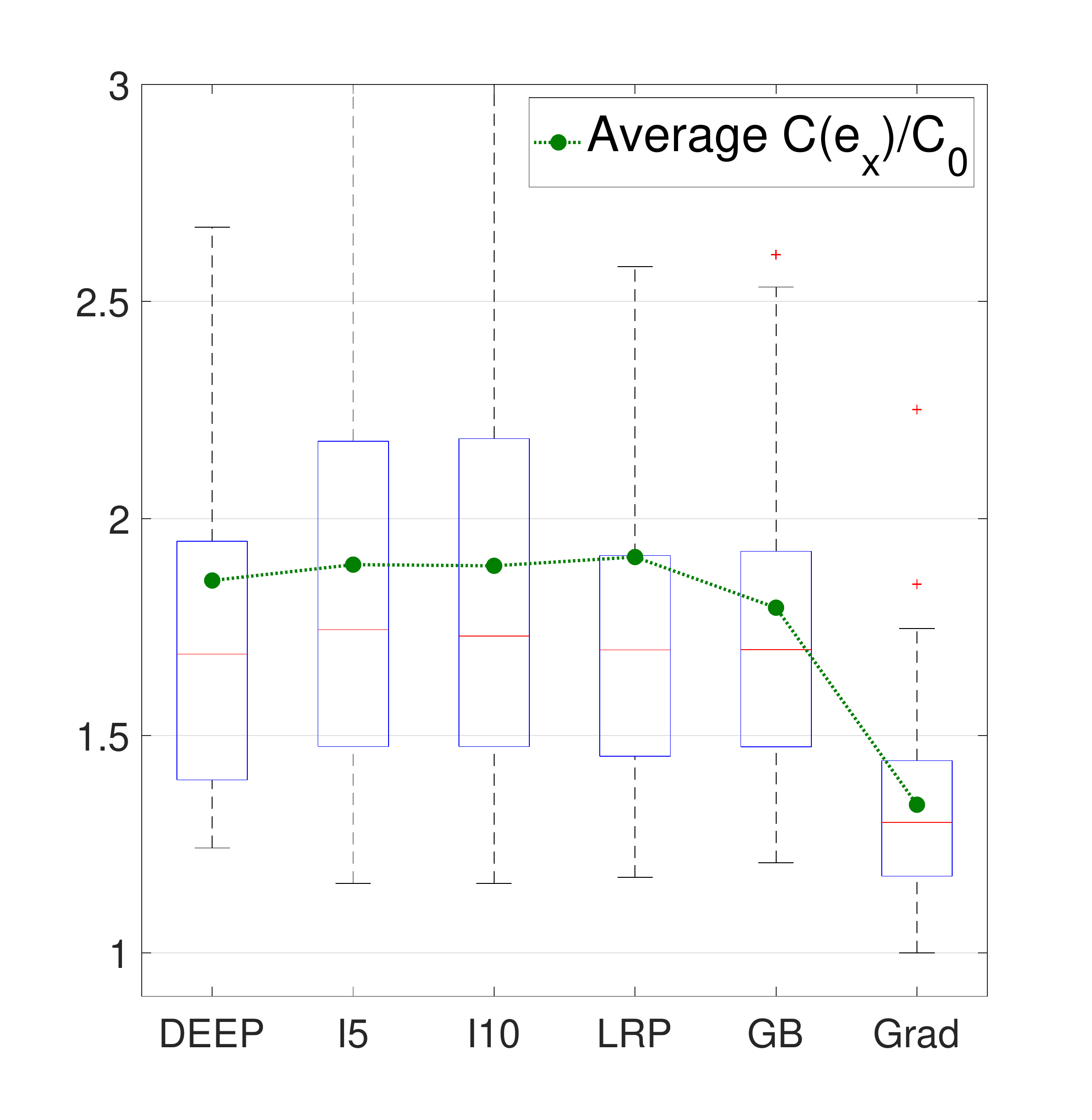}
			\caption{$c$-Eval of label 4 with GSA}
			\label{MNIST4G}
		\end{subfigure}
		\begin{subfigure}{.235\textwidth}
			\centering
			\includegraphics[scale=0.19]{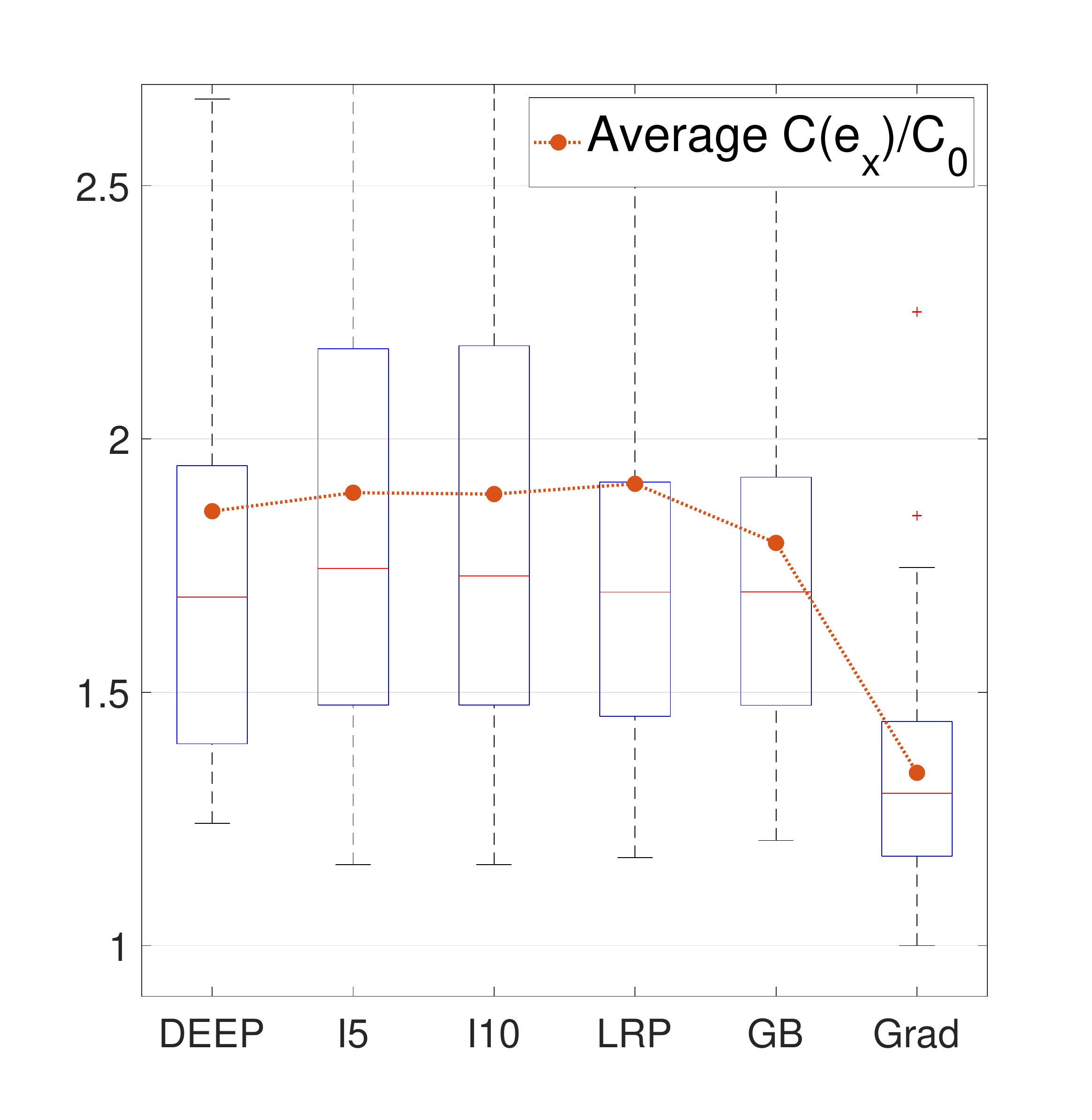}
			\caption{$c$-Eval of label 4 with IGA}
			\label{MNIST4I}
		\end{subfigure}
		
		%\hspace*{-2.0mm}%
		
		\begin{subfigure}{.235\textwidth}
			\centering
			\includegraphics[scale=0.19]{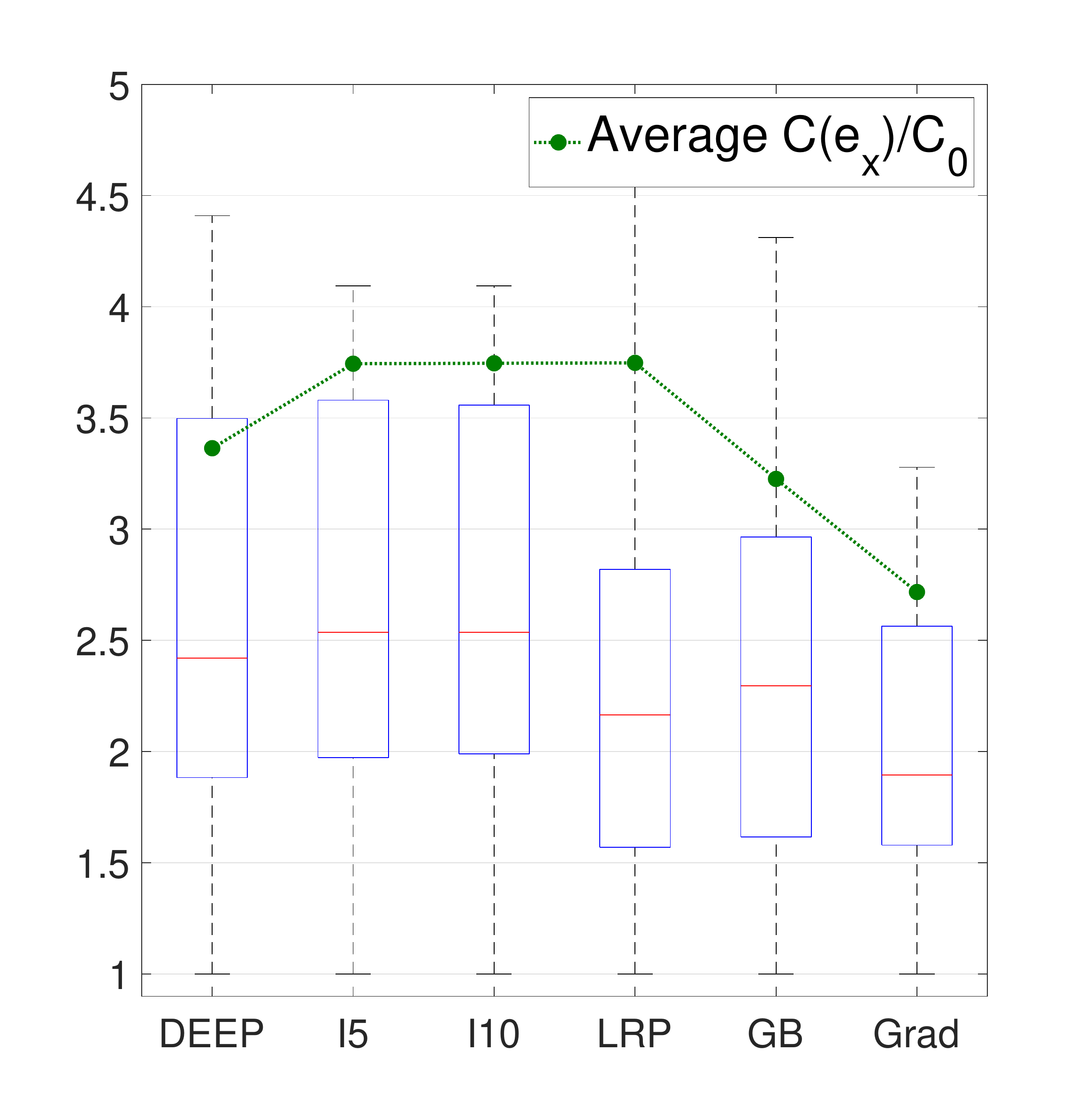}
			\caption{$c$-Eval of label 8 with GSA}
			\label{MNIST8G}
		\end{subfigure}
		\begin{subfigure}{.235\textwidth}
			\centering
			\includegraphics[scale=0.19]{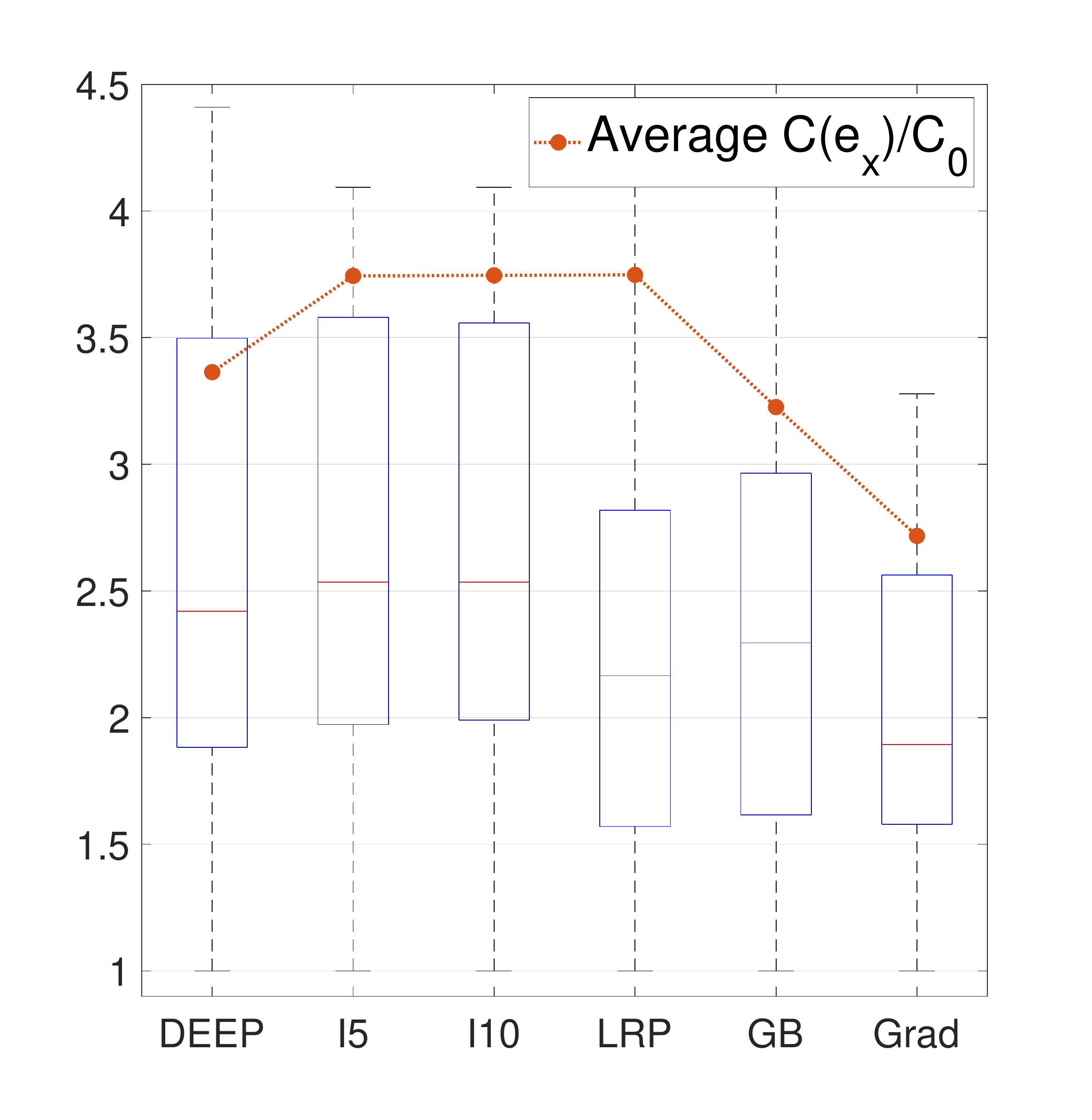}
			\caption{$c$-Eval of label 8 with IGA}
			\label{MNIST8I}
		\end{subfigure}
		
		%\hspace*{-2.0mm}%
		
		\begin{subfigure}{.235\textwidth}
			\centering
			\includegraphics[scale=0.19]{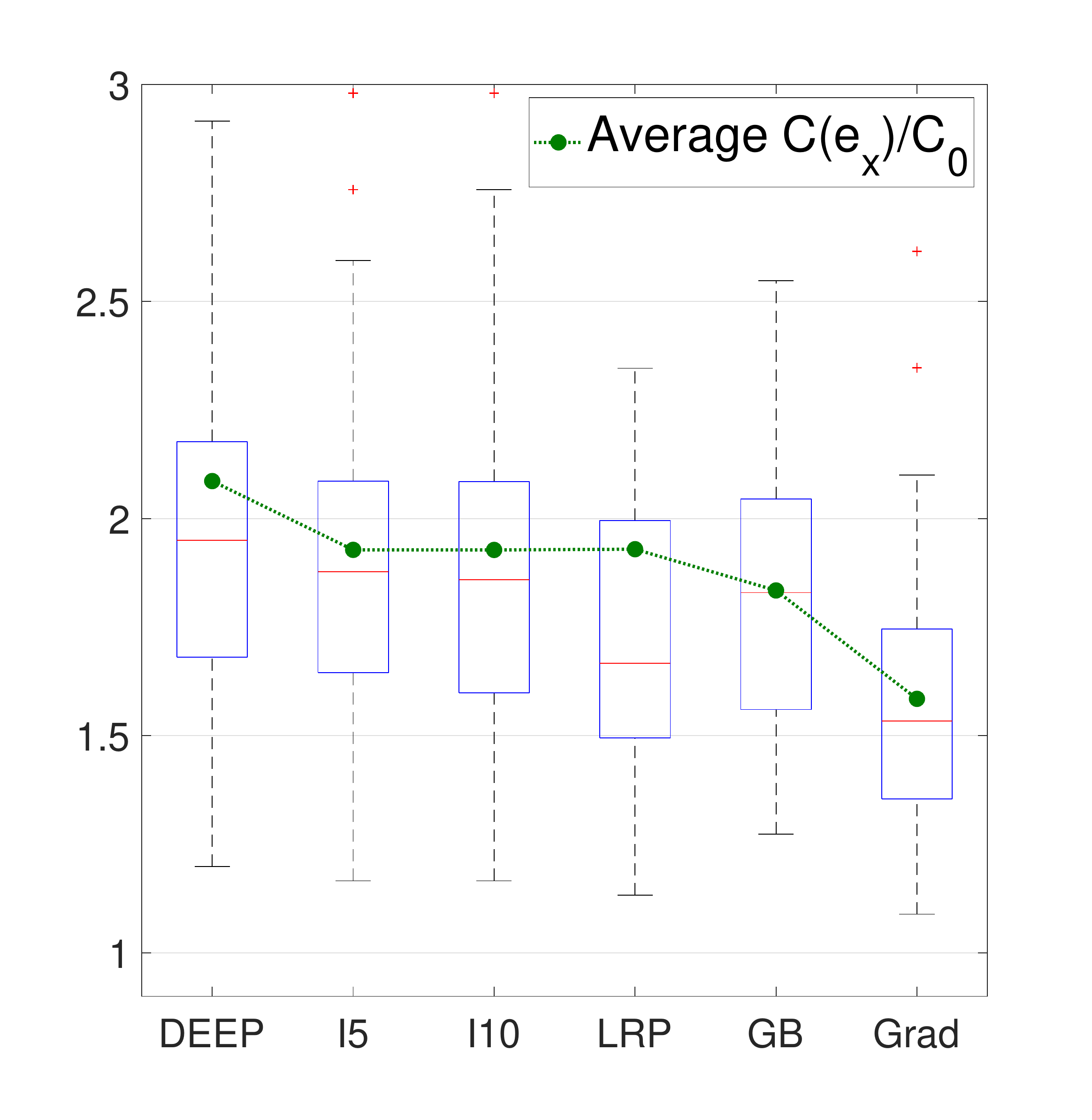}
			\caption{$c$-Eval of label 9 with GSA}
			\label{MNIST9G}
		\end{subfigure}
		\begin{subfigure}{.235\textwidth}
			\centering
			\includegraphics[scale=0.19]{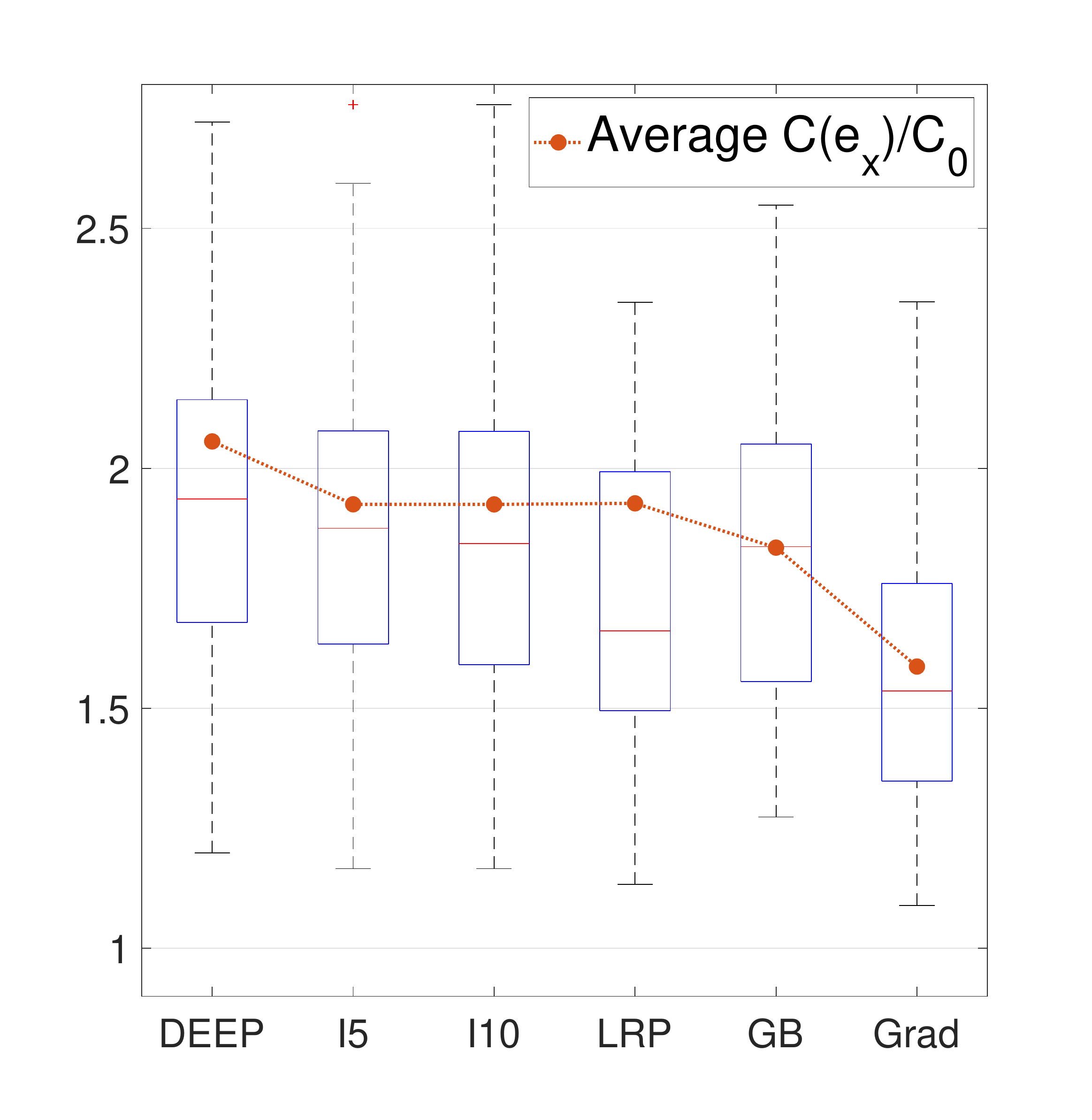}
			\caption{$c$-Eval of label 9 with IGA}
			\label{MNIST9I}
		\end{subfigure}
		
		\caption{We compute the $c$-Eval for 6 explainers on 1000 images of MNIST for labels $4, 8$ and $9$ to show the similarity between $c$-Eval and log-odds function in \cite{Avanti2017}.}
		\label{distno}
		%\vspace*{-3.0mm}
		\vspace{-4mm}
	\end{figure}
	
	To evaluate importance scores obtained by different methods on MNIST dataset, the authors of DeepLIFT designs the log-odds function as follows. Given an image that originally belongs to a class, they identify which pixels to erase to convert the original image to other target class and evaluate the change in the log-odds score between the two classes. The work conducted experiments of converting $8$ to $3$, $8$ to $6$, $9$ to $1$ and $4$ to $1$. In Fig.~\ref{distno}, we adopt $c$-Eval into the MNIST dataset to compare $c$-Eval of explainers with the corresponding log-odds scores. The figure displays the $c$-Eval of studied explainers on images with predictions $4, 8$ and $9$ respectively. We conduct the experiments using both GSA and IGA perturbation schemes. Besides the DeepLIFT in experiments for label 4 and 8, all relative ranking of explainers in $c$-Eval is consistent with the ranking resulted from log-odds computations shown in \cite{Avanti2017}. This result implies that our general frameworks of evaluating explainers based on $c$-Eval are applicable to this specific study on the MNIST dataset.

\subsection{Overall evaluations of explanations using c-Eval} \label{overall}
 Many interesting results and deductions can be drawn from experiments on MNIST, Caltech101 and CIFAR10. %We would like to integrate the experimental results and point out 
 We discuss several key observations in the followings.

Our first comment is about the correlation of $c$-Eval and the portion of predicted object captured by different explanations.
In CIFAR10 and especially Caltech101 (Fig.~\ref{examples_CIFAR} and~\ref{Caltechapp}, Appendix~\ref{exMC}), it is clear to us that most explanations containing the essential features of the predicted label have high $c$-Eval. %For MNIST, it is non-trivial to assess explainers' quality by a pure observation on Fig.~\ref{examples_MNIST} of Appendix~\ref{exMC}. This is also one of the main reasons motivating \cite{Avanti2017} to propose the log-odds function to evaluate explainers specifically for the MNIST dataset.

Our second attention is on the relative performance of GCam in three datasets. Since GCam is designed for convolutional neural networks such as the VGG19, we expect high relevant explanations from GCam in its experiments on Caltech101. However, as GCam exploits the last layer of the neural networks to generate the explanations~\cite{Ramprasaath2016}, we have low expectation on its capability of explaining predictions on MNIST and CIFAR10 dataset. The reason is that the models used in those two later dataset are too different from the VGG19. In fact, the adaptation of VGG19 on CIFAR10 \cite{Liu2015VeryDC} contains only 4 neurons in the last convolutional layer, which results in only 4 regions of the images that GCam can choose as region of high important (see explanations of Fig.~\ref{examples_CIFAR}, Appendix~\ref{exMC}). The distributions of $c$-Eval for two datasets in Fig.~\ref{examples_cifar10} and Fig.~\ref{Caltech101vggI} also reflect those expectations on GCam.

DeepLIFT is a back-propagation method and it is not only sensitive to the classifier structure but also the selection of reference image \cite{Avanti2017}. The experimental setups of DeepLIFT in the MNIST dataset shown in Fig.~\ref{dist} are taken directly from the source code of the explainer's paper. Our adoptions of DeepLIFT to CIFAR10 and Caltech101 are conducted without calibration on the reference image as the calibration procedure for color images is not provided. This might be the reason for the degradation of explainer's quality in these two datasets. It is clear that $c$-Eval captures this behavior.

\begin{figure}[h]
		\centering
		\begin{subfigure}{.1\textwidth}
			\centering
			\includegraphics[scale=1.8]{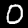}
			\captionsetup{labelformat=empty}
			\caption{Original}
		\end{subfigure}%
		\begin{subfigure}{.1\textwidth}
			\centering
			\includegraphics[scale=1.8]{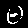}
			\captionsetup{labelformat=empty}
			\caption{1.97}
		\end{subfigure}%
		\begin{subfigure}{.1\textwidth}
			\centering
			\includegraphics[scale=1.8]{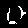}
			\captionsetup{labelformat=empty}
			\caption{1.96}
		\end{subfigure}%
		\begin{subfigure}{.1\textwidth}
			\centering
			\includegraphics[scale=1.8]{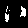}
			\captionsetup{labelformat=empty}
			\caption{1.41}
		\end{subfigure}%
		\begin{subfigure}{.1\textwidth}
			\centering
			\includegraphics[scale=1.8]{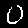}
			\captionsetup{labelformat=empty}
			\caption{2.01}
		\end{subfigure}%
		\begin{subfigure}{.1\textwidth}
			\centering
			\includegraphics[scale=1.8]{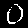}
			\captionsetup{labelformat=empty}
			\caption{1.99}
		\end{subfigure}%
		\begin{subfigure}{.1\textwidth}
			\centering
			\includegraphics[scale=1.8]{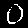}
			\captionsetup{labelformat=empty}
			\caption{1.99}
		\end{subfigure}%
		\begin{subfigure}{.1\textwidth}
			\centering
			\includegraphics[scale=1.8]{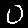}
			\captionsetup{labelformat=empty}
			\caption{1.94}
		\end{subfigure}%
		\begin{subfigure}{.1\textwidth}
			\centering
			\includegraphics[scale=1.8]{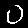}
			\captionsetup{labelformat=empty}
			\caption{2.05}
		\end{subfigure}%
		\begin{subfigure}{.1\textwidth}
			\centering
			\includegraphics[scale=1.8]{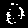}
			\captionsetup{labelformat=empty}
			\caption{2.02}
		\end{subfigure}%
		\\
		\begin{subfigure}{.1\textwidth}
			\centering
			\includegraphics[scale=1.8]{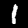}
			\captionsetup{labelformat=empty}
			\caption{Original}
		\end{subfigure}%
		\begin{subfigure}{.1\textwidth}
			\centering
			\includegraphics[scale=1.8]{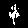}
			\captionsetup{labelformat=empty}
			\caption{1.93}
		\end{subfigure}%
		\begin{subfigure}{.1\textwidth}
			\centering
			\includegraphics[scale=1.8]{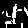}
			\captionsetup{labelformat=empty}
			\caption{1.62}
		\end{subfigure}%
		\begin{subfigure}{.1\textwidth}
			\centering
			\includegraphics[scale=1.8]{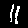}
			\captionsetup{labelformat=empty}
			\caption{1.67}
		\end{subfigure}%
		\begin{subfigure}{.1\textwidth}
			\centering
			\includegraphics[scale=1.8]{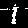}
			\captionsetup{labelformat=empty}
			\caption{1.40}
		\end{subfigure}%
		\begin{subfigure}{.1\textwidth}
			\centering
			\includegraphics[scale=1.8]{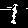}
			\captionsetup{labelformat=empty}
			\caption{1.44}
		\end{subfigure}%
		\begin{subfigure}{.1\textwidth}
			\centering
			\includegraphics[scale=1.8]{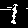}
			\captionsetup{labelformat=empty}
			\caption{1.44}
		\end{subfigure}%
		\begin{subfigure}{.1\textwidth}
			\centering
			\includegraphics[scale=1.8]{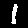}
			\captionsetup{labelformat=empty}
			\caption{1.44}
		\end{subfigure}%
		\begin{subfigure}{.1\textwidth}
			\centering
			\includegraphics[scale=1.8]{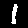}
			\captionsetup{labelformat=empty}
			\caption{1.44}
		\end{subfigure}%
		\begin{subfigure}{.1\textwidth}
			\centering
			\includegraphics[scale=1.8]{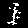}
			\captionsetup{labelformat=empty}
			\caption{1.34}
		\end{subfigure}%
		\\
		\begin{subfigure}{.1\textwidth}
			\centering
			\includegraphics[scale=1.8]{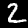}
			\captionsetup{labelformat=empty}
			\caption{Original}
		\end{subfigure}%
		\begin{subfigure}{.1\textwidth}
			\centering
			\includegraphics[scale=1.8]{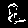}
			\captionsetup{labelformat=empty}
			\caption{2.92}
		\end{subfigure}%
		\begin{subfigure}{.1\textwidth}
			\centering
			\includegraphics[scale=1.8]{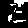}
			\captionsetup{labelformat=empty}
			\caption{2.72}
		\end{subfigure}%
		\begin{subfigure}{.1\textwidth}
			\centering
			\includegraphics[scale=1.8]{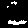}
			\captionsetup{labelformat=empty}
			\caption{1.09}
		\end{subfigure}%
		\begin{subfigure}{.1\textwidth}
			\centering
			\includegraphics[scale=1.8]{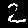}
			\captionsetup{labelformat=empty}
			\caption{2.78}
		\end{subfigure}%
		\begin{subfigure}{.1\textwidth}
			\centering
			\includegraphics[scale=1.8]{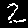}
			\captionsetup{labelformat=empty}
			\caption{3.10}
		\end{subfigure}%
		\begin{subfigure}{.1\textwidth}
			\centering
			\includegraphics[scale=1.8]{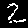}
			\captionsetup{labelformat=empty}
			\caption{3.10}
		\end{subfigure}%
		\begin{subfigure}{.1\textwidth}
			\centering
			\includegraphics[scale=1.8]{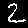}
			\captionsetup{labelformat=empty}
			\caption{2.64}
		\end{subfigure}%
		\begin{subfigure}{.1\textwidth}
			\centering
			\includegraphics[scale=1.8]{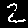}
			\captionsetup{labelformat=empty}
			\caption{2.68}
		\end{subfigure}%
		\begin{subfigure}{.1\textwidth}
			\centering
			\includegraphics[scale=1.8]{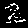}
			\captionsetup{labelformat=empty}
			\caption{2.82}
		\end{subfigure}%
		\\
		\begin{subfigure}{.1\textwidth}
			\centering
			\includegraphics[scale=1.8]{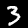}
			\captionsetup{labelformat=empty}
			\caption{Original}
		\end{subfigure}%
		\begin{subfigure}{.1\textwidth}
			\centering
			\includegraphics[scale=1.8]{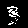}
			\captionsetup{labelformat=empty}
			\caption{1.52}
		\end{subfigure}%
		\begin{subfigure}{.1\textwidth}
			\centering
			\includegraphics[scale=1.8]{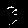}
			\captionsetup{labelformat=empty}
			\caption{1.22}
		\end{subfigure}%
		\begin{subfigure}{.1\textwidth}
			\centering
			\includegraphics[scale=1.8]{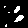}
			\captionsetup{labelformat=empty}
			\caption{2.01}
		\end{subfigure}%
		\begin{subfigure}{.1\textwidth}
			\centering
			\includegraphics[scale=1.8]{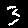}
			\captionsetup{labelformat=empty}
			\caption{1.29}
		\end{subfigure}%
		\begin{subfigure}{.1\textwidth}
			\centering
			\includegraphics[scale=1.8]{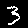}
			\captionsetup{labelformat=empty}
			\caption{1.27}
		\end{subfigure}%
		\begin{subfigure}{.1\textwidth}
			\centering
			\includegraphics[scale=1.8]{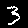}
			\captionsetup{labelformat=empty}
			\caption{1.27}
		\end{subfigure}%
		\begin{subfigure}{.1\textwidth}
			\centering
			\includegraphics[scale=1.8]{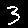}
			\captionsetup{labelformat=empty}
			\caption{1.29}
		\end{subfigure}%
		\begin{subfigure}{.1\textwidth}
			\centering
			\includegraphics[scale=1.8]{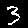}
			\captionsetup{labelformat=empty}
			\caption{1.29}
		\end{subfigure}%
		\begin{subfigure}{.1\textwidth}
			\centering
			\includegraphics[scale=1.8]{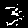}
			\captionsetup{labelformat=empty}
			\caption{1.26}
		\end{subfigure}%
		\\
		\begin{subfigure}{.1\textwidth}
			\centering
			\includegraphics[scale=1.8]{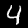}
			\captionsetup{labelformat=empty}
			\caption{Original}
		\end{subfigure}%
		\begin{subfigure}{.1\textwidth}
			\centering
			\includegraphics[scale=1.8]{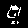}
			\captionsetup{labelformat=empty}
			\caption{1.51}
		\end{subfigure}%
		\begin{subfigure}{.1\textwidth}
			\centering
			\includegraphics[scale=1.8]{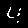}
			\captionsetup{labelformat=empty}
			\caption{1.07}
		\end{subfigure}%
		\begin{subfigure}{.1\textwidth}
			\centering
			\includegraphics[scale=1.8]{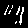}
			\captionsetup{labelformat=empty}
			\caption{1.22}
		\end{subfigure}%
		\begin{subfigure}{.1\textwidth}
			\centering
			\includegraphics[scale=1.8]{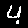}
			\captionsetup{labelformat=empty}
			\caption{1.19}
		\end{subfigure}%
		\begin{subfigure}{.1\textwidth}
			\centering
			\includegraphics[scale=1.8]{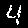}
			\captionsetup{labelformat=empty}
			\caption{1.24}
		\end{subfigure}%
		\begin{subfigure}{.1\textwidth}
			\centering
			\includegraphics[scale=1.8]{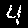}
			\captionsetup{labelformat=empty}
			\caption{1.24}
		\end{subfigure}%
		\begin{subfigure}{.1\textwidth}
			\centering
			\includegraphics[scale=1.8]{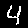}
			\captionsetup{labelformat=empty}
			\caption{1.19}
		\end{subfigure}%
		\begin{subfigure}{.1\textwidth}
			\centering
			\includegraphics[scale=1.8]{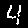}
			\captionsetup{labelformat=empty}
			\caption{1.25}
		\end{subfigure}%
		\begin{subfigure}{.1\textwidth}
			\centering
			\includegraphics[scale=1.8]{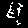}
			\captionsetup{labelformat=empty}
			\caption{1.18}
		\end{subfigure}%
		\\
		\begin{subfigure}{.1\textwidth}
			\centering
			\includegraphics[scale=1.8]{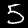}
			\captionsetup{labelformat=empty}
			\caption{Original}
		\end{subfigure}%
		\begin{subfigure}{.1\textwidth}
			\centering
			\includegraphics[scale=1.8]{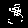}
			\captionsetup{labelformat=empty}
			\caption{3.94}
		\end{subfigure}%
		\begin{subfigure}{.1\textwidth}
			\centering
			\includegraphics[scale=1.8]{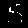}
			\captionsetup{labelformat=empty}
			\caption{2.76}
		\end{subfigure}%
		\begin{subfigure}{.1\textwidth}
			\centering
			\includegraphics[scale=1.8]{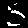}
			\captionsetup{labelformat=empty}
			\caption{1.95}
		\end{subfigure}%
		\begin{subfigure}{.1\textwidth}
			\centering
			\includegraphics[scale=1.8]{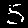}
			\captionsetup{labelformat=empty}
			\caption{2.86}
		\end{subfigure}%
		\begin{subfigure}{.1\textwidth}
			\centering
			\includegraphics[scale=1.8]{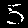}
			\captionsetup{labelformat=empty}
			\caption{2.97}
		\end{subfigure}%
		\begin{subfigure}{.1\textwidth}
			\centering
			\includegraphics[scale=1.8]{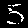}
			\captionsetup{labelformat=empty}
			\caption{2.97}
		\end{subfigure}%
		\begin{subfigure}{.1\textwidth}
			\centering
			\includegraphics[scale=1.8]{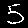}
			\captionsetup{labelformat=empty}
			\caption{2.47}
		\end{subfigure}%
		\begin{subfigure}{.1\textwidth}
			\centering
			\includegraphics[scale=1.8]{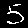}
			\captionsetup{labelformat=empty}
			\caption{2.47}
		\end{subfigure}%
		\begin{subfigure}{.1\textwidth}
			\centering
			\includegraphics[scale=1.8]{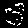}
			\captionsetup{labelformat=empty}
			\caption{2.81}
		\end{subfigure}%
		\\
		\begin{subfigure}{.1\textwidth}
			\centering
			\includegraphics[scale=1.8]{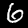}
			\captionsetup{labelformat=empty}
			\caption{Original}
		\end{subfigure}%
		\begin{subfigure}{.1\textwidth}
			\centering
			\includegraphics[scale=1.8]{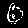}
			\captionsetup{labelformat=empty}
			\caption{1.84}
		\end{subfigure}%
		\begin{subfigure}{.1\textwidth}
			\centering
			\includegraphics[scale=1.8]{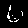}
			\captionsetup{labelformat=empty}
			\caption{1.14}
		\end{subfigure}%
		\begin{subfigure}{.1\textwidth}
			\centering
			\includegraphics[scale=1.8]{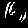}
			\captionsetup{labelformat=empty}
			\caption{1.10}
		\end{subfigure}%
		\begin{subfigure}{.1\textwidth}
			\centering
			\includegraphics[scale=1.8]{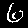}
			\captionsetup{labelformat=empty}
			\caption{1.34}
		\end{subfigure}%
		\begin{subfigure}{.1\textwidth}
			\centering
			\includegraphics[scale=1.8]{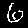}
			\captionsetup{labelformat=empty}
			\caption{1.41}
		\end{subfigure}%
		\begin{subfigure}{.1\textwidth}
			\centering
			\includegraphics[scale=1.8]{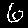}
			\captionsetup{labelformat=empty}
			\caption{1.41}
		\end{subfigure}%
		\begin{subfigure}{.1\textwidth}
			\centering
			\includegraphics[scale=1.8]{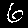}
			\captionsetup{labelformat=empty}
			\caption{1.37}
		\end{subfigure}%
		\begin{subfigure}{.1\textwidth}
			\centering
			\includegraphics[scale=1.8]{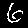}
			\captionsetup{labelformat=empty}
			\caption{1.40}
		\end{subfigure}%
		\begin{subfigure}{.1\textwidth}
			\centering
			\includegraphics[scale=1.8]{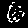}
			\captionsetup{labelformat=empty}
			\caption{1.28}
		\end{subfigure}%
		\\
		\begin{subfigure}{.1\textwidth}
			\centering
			\includegraphics[scale=1.8]{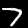}
			\captionsetup{labelformat=empty}
			\caption{Original}
		\end{subfigure}%
		\begin{subfigure}{.1\textwidth}
			\centering
			\includegraphics[scale=1.8]{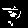}
			\captionsetup{labelformat=empty}
			\caption{1.55}
		\end{subfigure}%
		\begin{subfigure}{.1\textwidth}
			\centering
			\includegraphics[scale=1.8]{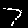}
			\captionsetup{labelformat=empty}
			\caption{1.26}
		\end{subfigure}%
		\begin{subfigure}{.1\textwidth}
			\centering
			\includegraphics[scale=1.8]{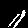}
			\captionsetup{labelformat=empty}
			\caption{1.32}
		\end{subfigure}%
		\begin{subfigure}{.1\textwidth}
			\centering
			\includegraphics[scale=1.8]{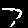}
			\captionsetup{labelformat=empty}
			\caption{1.36}
		\end{subfigure}%
		\begin{subfigure}{.1\textwidth}
			\centering
			\includegraphics[scale=1.8]{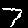}
			\captionsetup{labelformat=empty}
			\caption{1.37}
		\end{subfigure}%
		\begin{subfigure}{.1\textwidth}
			\centering
			\includegraphics[scale=1.8]{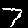}
			\captionsetup{labelformat=empty}
			\caption{1.37}
		\end{subfigure}%
		\begin{subfigure}{.1\textwidth}
			\centering
			\includegraphics[scale=1.8]{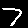}
			\captionsetup{labelformat=empty}
			\caption{1.35}
		\end{subfigure}%
		\begin{subfigure}{.1\textwidth}
			\centering
			\includegraphics[scale=1.8]{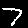}
			\captionsetup{labelformat=empty}
			\caption{1.33}
		\end{subfigure}%
		\begin{subfigure}{.1\textwidth}
			\centering
			\includegraphics[scale=1.8]{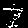}
			\captionsetup{labelformat=empty}
			\caption{1.32}
		\end{subfigure}%
		\\
		\begin{subfigure}{.1\textwidth}
			\centering
			\includegraphics[scale=1.8]{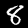}
			\captionsetup{labelformat=empty}
			\caption{Original}
		\end{subfigure}%
		\begin{subfigure}{.1\textwidth}
			\centering
			\includegraphics[scale=1.8]{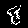}
			\captionsetup{labelformat=empty}
			\caption{2.53}
		\end{subfigure}%
		\begin{subfigure}{.1\textwidth}
			\centering
			\includegraphics[scale=1.8]{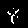}
			\captionsetup{labelformat=empty}
			\caption{1.98}
		\end{subfigure}%
		\begin{subfigure}{.1\textwidth}
			\centering
			\includegraphics[scale=1.8]{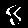}
			\captionsetup{labelformat=empty}
			\caption{1.76}
		\end{subfigure}%
		\begin{subfigure}{.1\textwidth}
			\centering
			\includegraphics[scale=1.8]{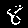}
			\captionsetup{labelformat=empty}
			\caption{2.49}
		\end{subfigure}%
		\begin{subfigure}{.1\textwidth}
			\centering
			\includegraphics[scale=1.8]{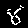}
			\captionsetup{labelformat=empty}
			\caption{2.56}
		\end{subfigure}%
		\begin{subfigure}{.1\textwidth}
			\centering
			\includegraphics[scale=1.8]{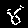}
			\captionsetup{labelformat=empty}
			\caption{2.56}
		\end{subfigure}%
		\begin{subfigure}{.1\textwidth}
			\centering
			\includegraphics[scale=1.8]{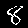}
			\captionsetup{labelformat=empty}
			\caption{2.37}
		\end{subfigure}%
		\begin{subfigure}{.1\textwidth}
			\centering
			\includegraphics[scale=1.8]{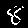}
			\captionsetup{labelformat=empty}
			\caption{2.37}
		\end{subfigure}%
		\begin{subfigure}{.1\textwidth}
			\centering
			\includegraphics[scale=1.8]{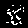}
			\captionsetup{labelformat=empty}
			\caption{2.29}
		\end{subfigure}%
		\\
		\begin{subfigure}{.1\textwidth}
			\centering
			\includegraphics[scale=1.8]{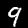}
			\captionsetup{labelformat=empty}
			\caption{Original}
		\end{subfigure}%
		\begin{subfigure}{.1\textwidth}
			\centering
			\includegraphics[scale=1.8]{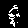}
			\captionsetup{labelformat=empty}
			\caption{1.72}
		\end{subfigure}%
		\begin{subfigure}{.1\textwidth}
			\centering
			\includegraphics[scale=1.8]{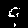}
			\captionsetup{labelformat=empty}
			\caption{1.55}
		\end{subfigure}%
		\begin{subfigure}{.1\textwidth}
			\centering
			\includegraphics[scale=1.8]{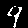}
			\captionsetup{labelformat=empty}
			\caption{1.18}
		\end{subfigure}%
		\begin{subfigure}{.1\textwidth}
			\centering
			\includegraphics[scale=1.8]{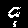}
			\captionsetup{labelformat=empty}
			\caption{1.97}
		\end{subfigure}%
		\begin{subfigure}{.1\textwidth}
			\centering
			\includegraphics[scale=1.8]{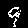}
			\captionsetup{labelformat=empty}
			\caption{1.95}
		\end{subfigure}%
		\begin{subfigure}{.1\textwidth}
			\centering
			\includegraphics[scale=1.8]{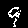}
			\captionsetup{labelformat=empty}
			\caption{1.95}
		\end{subfigure}%
		\begin{subfigure}{.1\textwidth}
			\centering
			\includegraphics[scale=1.8]{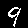}
			\captionsetup{labelformat=empty}
			\caption{1.93}
		\end{subfigure}%
		\begin{subfigure}{.1\textwidth}
			\centering
			\includegraphics[scale=1.8]{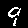}
			\captionsetup{labelformat=empty}
			\caption{1.93}
		\end{subfigure}%
		\begin{subfigure}{.1\textwidth}
			\centering
			\includegraphics[scale=1.8]{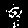}
			\captionsetup{labelformat=empty}
			\caption{1.89}
		\end{subfigure}%

		\caption{Some examples of explanations and $c$-Eval on MNIST. The explainers from left to right: SHAP, LIME, GCam, DeepLIFT, Integrated Gradient with 5 and 10 interpolations, Guided Backpropagation, and Gradient. The number associated with each figure is the ratio $c_{f, \bm x}(e_{\bm x}) / c_{f, \bm x}(\emptyset)$. It is non-intuitive to evaluate these explanations purely by observations.}
		\label{examples_MNIST}
	\end{figure}

Our final remark is on the exceptionally high $c$-Eval of SHAP shown in all three datasets. This result encourages us to take a deeper look at explanations produced by SHAP. A quick glance at SHAP on MNIST in 
 Fig.~\ref{examples_MNIST} (Appendix~\ref{exMC}) might suggest that the explainer is worse than some other back-propagation methods such as DeepLIFT, Integrated Gradient or GB; however, the figure shows that SHAP captures some important features that are overlooked by others. Let's consider the explanation of number 4 as an example. SHAP is the only explainer detecting that the black area on top of number 4 is important. In fact, this area is essential to the prediction since, if these pixels are white instead of black, the original prediction should be 0 instead of 4. Without the $c$-Eval computations, we might solely assess SHAP based on intuitive observations and wrongly evaluate the explainer.

\section{Conclusions and Future Works}\label{conclusion}
In this paper, we introduce $c$-Eval to evaluate explanations of various feature-based explainers. Extensive experiments show that $c$-Eval of explanation reflects the importance of features included in the explanation. This study leads to several interesting research questions for the future work. For example, the distributions of $c$-Eval in Fig.~\ref{dist} advocates that there is a fundamental difference between the quality of black-box explainers (SHAP, LIME and GCam) and back-propagation explainers (DEEP, Integrated Gradients, LRP, GB and Grad), which is ambiguous prior to this work. %On the other hand, one of our interest is to solve for the explanation that maximize $c$-Eval. 
From the novelty of $c$-Eval, we expect that knowledge on the explanation maximizing $c$-Eval will offer us a much clearer view on predictions made by modern neural networks. 

 %$c$-Eval offers us a clear quantification that might shed light into many unanswered questions behind machine learning explainers. 

\bibliographystyle{IEEEtran}
\bibliography{minhbibfile}
\newpage
\pagebreak
\onecolumn
\appendices

	\section{Experiments on CIFAR10} \label{CIFAR10app}
	
\begin{figure}[h]
		\centering
		\includegraphics[scale=0.3]{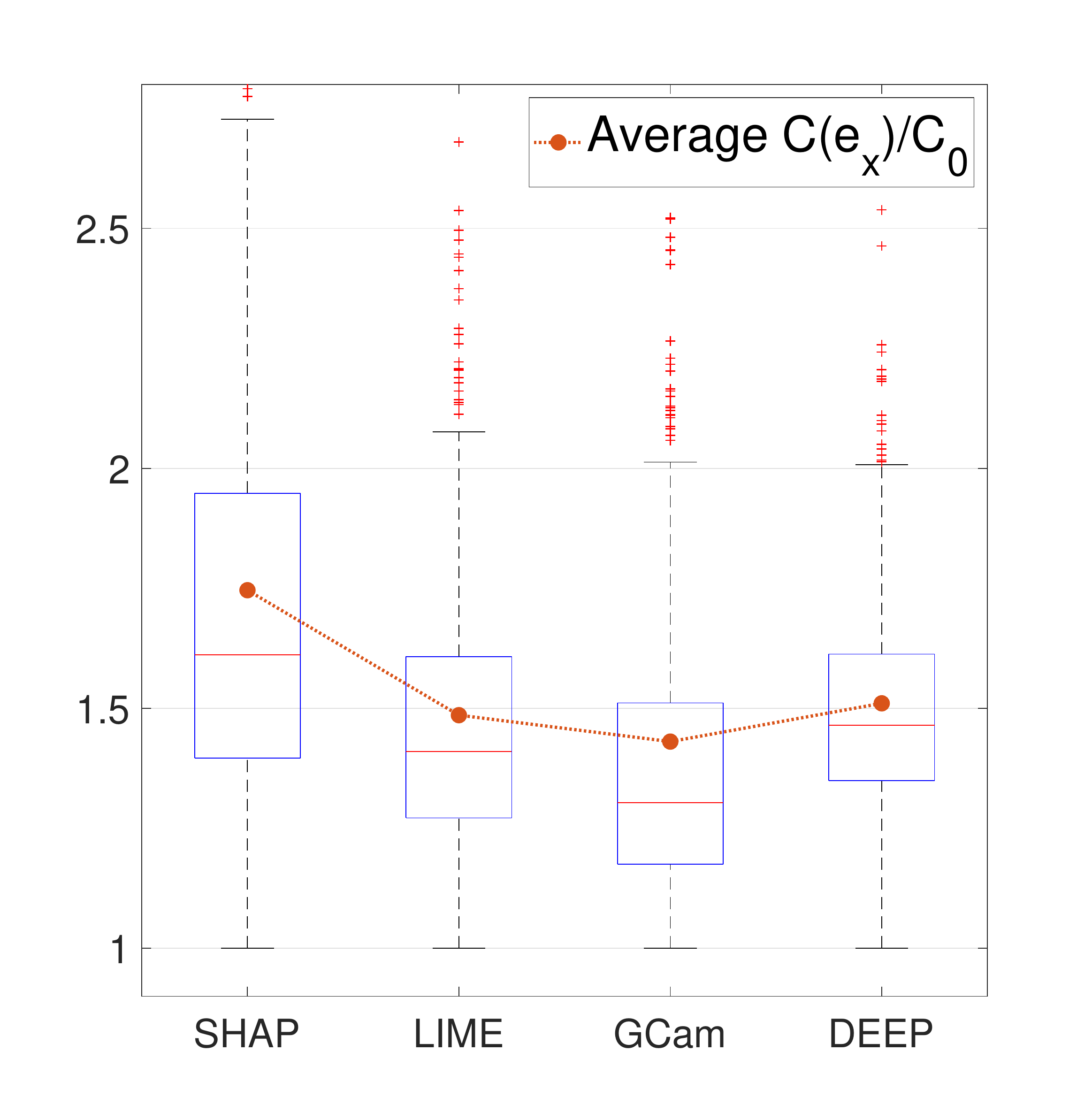}
		\caption{Distributions of $c$-Eval on CIFAR10 dataset.}
		\label{examples_cifar10}
	\end{figure}

\begin{figure}[h]
		\centering
		\begin{subfigure}{.2\textwidth}
			\centering
			\includegraphics[scale=0.232]{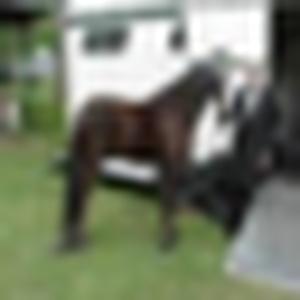}
			\captionsetup{labelformat=empty}
			\caption{Original}
		\end{subfigure}%
		\begin{subfigure}{.2\textwidth}
			\centering
			\includegraphics[scale=0.232]{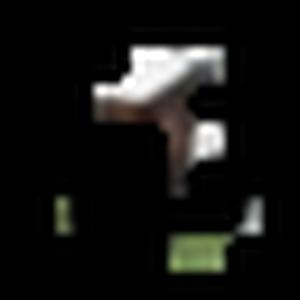}
			\captionsetup{labelformat=empty}
			\caption{1.66}
		\end{subfigure}%
		\begin{subfigure}{.2\textwidth}
			\centering
			\includegraphics[scale=0.232]{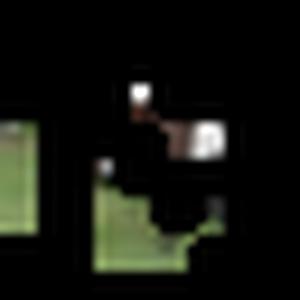}
			\captionsetup{labelformat=empty}
			\caption{1.32}
		\end{subfigure}%
		\begin{subfigure}{.2\textwidth}
			\centering
			\includegraphics[scale=0.232]{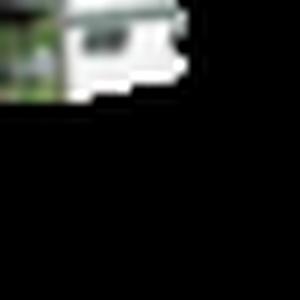}
			\captionsetup{labelformat=empty}
			\caption{1.03}
		\end{subfigure}%
		\begin{subfigure}{.2\textwidth}
			\centering
			\includegraphics[scale=0.232]{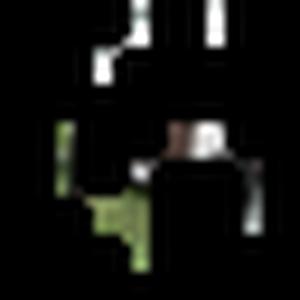}
			\captionsetup{labelformat=empty}
			\caption{1.41}
		\end{subfigure}%
		\\
		\begin{subfigure}{.2\textwidth}
			\centering
			\includegraphics[scale=0.232]{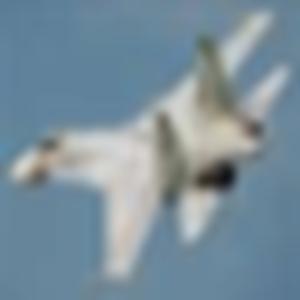}
			\captionsetup{labelformat=empty}
			\caption{Original}
		\end{subfigure}%
		\begin{subfigure}{.2\textwidth}
			\centering
			\includegraphics[scale=0.232]{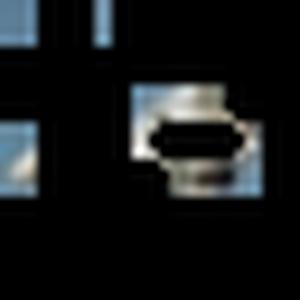}
			\captionsetup{labelformat=empty}
			\caption{2.11}
		\end{subfigure}%
		\begin{subfigure}{.2\textwidth}
			\centering
			\includegraphics[scale=0.232]{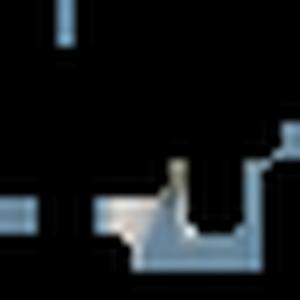}
			\captionsetup{labelformat=empty}
			\caption{1.81}
		\end{subfigure}%
		\begin{subfigure}{.2\textwidth}
			\centering
			\includegraphics[scale=0.232]{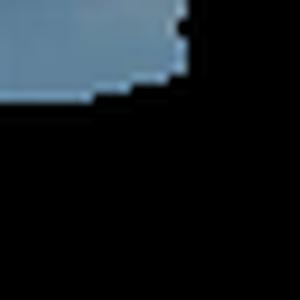}
			\captionsetup{labelformat=empty}
			\caption{1.31}
		\end{subfigure}%
		\begin{subfigure}{.2\textwidth}
			\centering
			\includegraphics[scale=0.232]{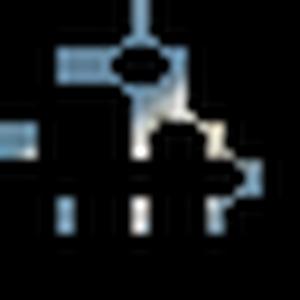}
			\captionsetup{labelformat=empty}
			\caption{1.32}
		\end{subfigure}%
		\\
		\begin{subfigure}{.2\textwidth}
			\centering
			\includegraphics[scale=0.232]{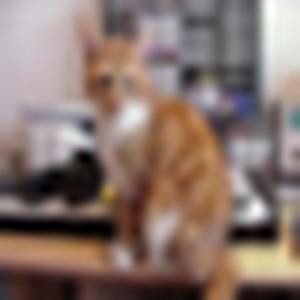}
			\captionsetup{labelformat=empty}
			\caption{Original}
		\end{subfigure}%
		\begin{subfigure}{.2\textwidth}
			\centering
			\includegraphics[scale=0.232]{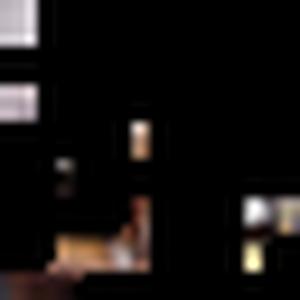}
			\captionsetup{labelformat=empty}
			\caption{1.19}
		\end{subfigure}%
		\begin{subfigure}{.2\textwidth}
			\centering
			\includegraphics[scale=0.232]{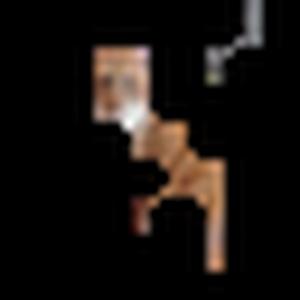}
			\captionsetup{labelformat=empty}
			\caption{1.29}
		\end{subfigure}%
		\begin{subfigure}{.2\textwidth}
			\centering
			\includegraphics[scale=0.232]{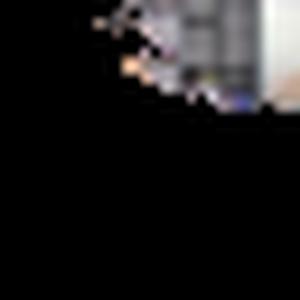}
			\captionsetup{labelformat=empty}
			\caption{1.06}
		\end{subfigure}%
		\begin{subfigure}{.2\textwidth}
			\centering
			\includegraphics[scale=0.232]{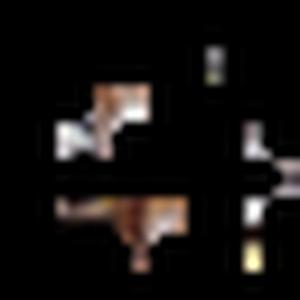}
			\captionsetup{labelformat=empty}
			\caption{1.51}
		\end{subfigure}%
		\\
		\begin{subfigure}{.2\textwidth}
			\centering
			\includegraphics[scale=0.232]{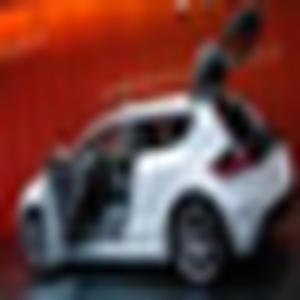}
			\captionsetup{labelformat=empty}
			\caption{Original}
		\end{subfigure}%
		\begin{subfigure}{.2\textwidth}
			\centering
			\includegraphics[scale=0.232]{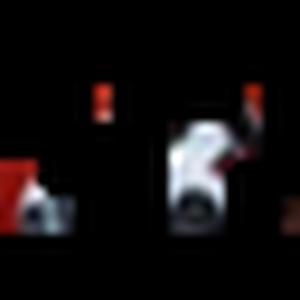}
			\captionsetup{labelformat=empty}
			\caption{1.91}
		\end{subfigure}%
		\begin{subfigure}{.2\textwidth}
			\centering
			\includegraphics[scale=0.232]{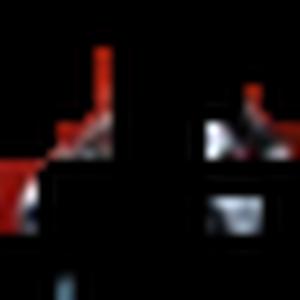}
			\captionsetup{labelformat=empty}
			\caption{1.75}
		\end{subfigure}%
		\begin{subfigure}{.2\textwidth}
			\centering
			\includegraphics[scale=0.232]{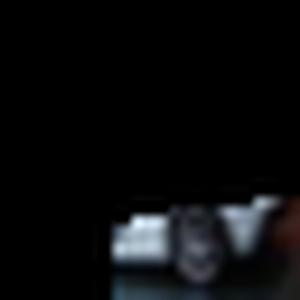}
			\captionsetup{labelformat=empty}
			\caption{2.00}
		\end{subfigure}%
		\begin{subfigure}{.2\textwidth}
			\centering
			\includegraphics[scale=0.232]{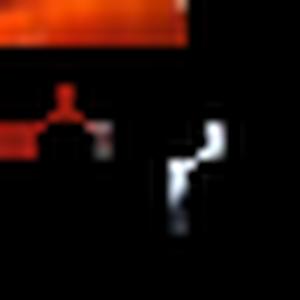}
			\captionsetup{labelformat=empty}
			\caption{1.31}
		\end{subfigure}%
		\\
		\begin{subfigure}{.2\textwidth}
			\centering
			\includegraphics[scale=0.232]{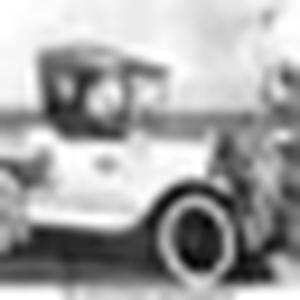}
			\captionsetup{labelformat=empty}
			\caption{Original}
		\end{subfigure}%
		\begin{subfigure}{.2\textwidth}
			\centering
			\includegraphics[scale=0.232]{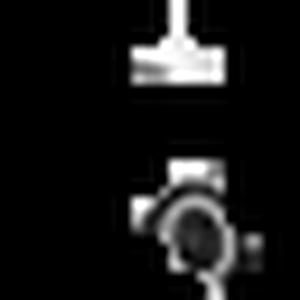}
			\captionsetup{labelformat=empty}
			\caption{1.85}
		\end{subfigure}%
		\begin{subfigure}{.2\textwidth}
			\centering
			\includegraphics[scale=0.232]{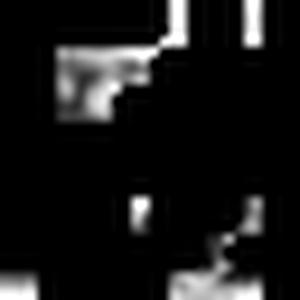}
			\captionsetup{labelformat=empty}
			\caption{1.60}
		\end{subfigure}%
		\begin{subfigure}{.2\textwidth}
			\centering
			\includegraphics[scale=0.232]{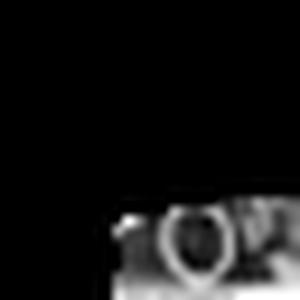}
			\captionsetup{labelformat=empty}
			\caption{2.52}
		\end{subfigure}%
		\begin{subfigure}{.2\textwidth}
			\centering
			\includegraphics[scale=0.232]{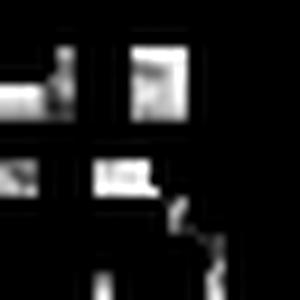}
			\captionsetup{labelformat=empty}
			\caption{1.38}
		\end{subfigure}%
		\\
		\begin{subfigure}{.2\textwidth}
			\centering
			\includegraphics[scale=0.232]{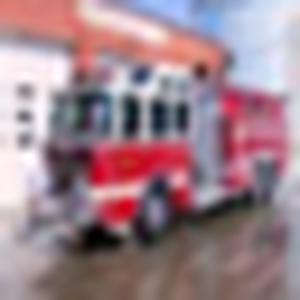}
			\captionsetup{labelformat=empty}
			\caption{Original}
		\end{subfigure}%
		\begin{subfigure}{.2\textwidth}
			\centering
			\includegraphics[scale=0.232]{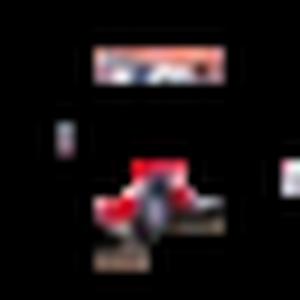}
			\captionsetup{labelformat=empty}
			\caption{2.77}
		\end{subfigure}%
		\begin{subfigure}{.2\textwidth}
			\centering
			\includegraphics[scale=0.232]{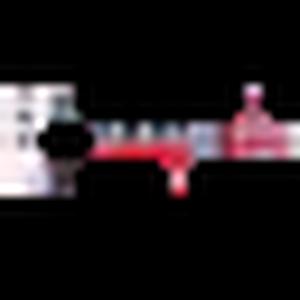}
			\captionsetup{labelformat=empty}
			\caption{1.32}
		\end{subfigure}%
		\begin{subfigure}{.2\textwidth}
			\centering
			\includegraphics[scale=0.232]{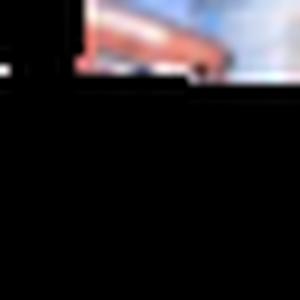}
			\captionsetup{labelformat=empty}
			\caption{1.11}
		\end{subfigure}%
		\begin{subfigure}{.2\textwidth}
			\centering
			\includegraphics[scale=0.232]{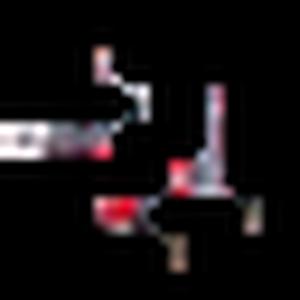}
			\captionsetup{labelformat=empty}
			\caption{2.19}
		\end{subfigure}%
		\\
		\begin{subfigure}{.2\textwidth}
			\centering
			\includegraphics[scale=0.232]{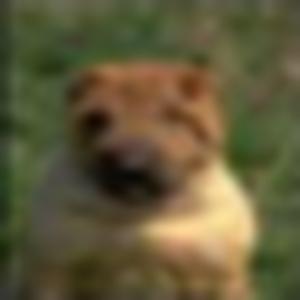}
			\captionsetup{labelformat=empty}
			\caption{Original}
		\end{subfigure}%
		\begin{subfigure}{.2\textwidth}
			\centering
			\includegraphics[scale=0.232]{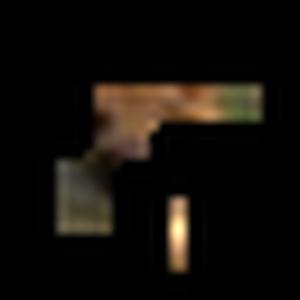}
			\captionsetup{labelformat=empty}
			\caption{1.75}
		\end{subfigure}%
		\begin{subfigure}{.2\textwidth}
			\centering
			\includegraphics[scale=0.232]{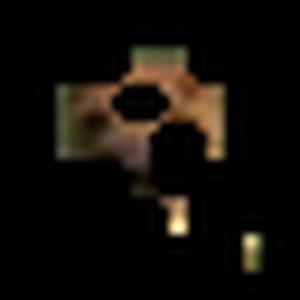}
			\captionsetup{labelformat=empty}
			\caption{1.72}
		\end{subfigure}%
		\begin{subfigure}{.2\textwidth}
			\centering
			\includegraphics[scale=0.232]{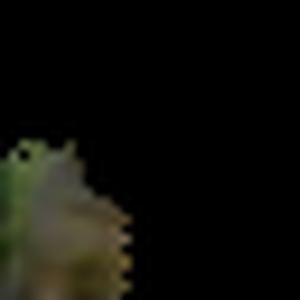}
			\captionsetup{labelformat=empty}
			\caption{1.13}
		\end{subfigure}%
		\begin{subfigure}{.2\textwidth}
			\centering
			\includegraphics[scale=0.232]{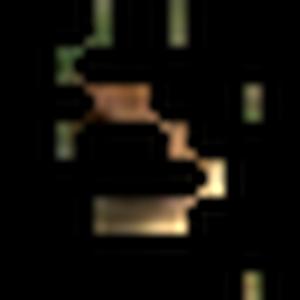}
			\captionsetup{labelformat=empty}
			\caption{1.43}
		\end{subfigure}%
		
		\caption{Some examples of Explanations and $c$-Eval on CIFAR10. The explainers from left to right: SHAP, LIME, GCam and DeepLIFT. The number associated with each figure is the ratio $c_{f, \bm x}(e_{\bm x}) / c_{f, \bm x}(\emptyset)$. We observe that most explanations which capture the signature components of the images have relatively high $c$-Eval.}
		\label{examples_CIFAR}
	\end{figure}

	Besides MNIST and Caltech101, we also conduct experiment on the small color image dataset CIFAR10 \cite{Krizhevsky2009}. The distributions of $c$-Eval on 500 images of the dataset and some examples are shown in Figs.~\ref{examples_cifar10} and~\ref{examples_CIFAR}. The experimental parameters model and the segmentation procedure are similar to that on Caltech101 dataset. The classifier model in this experiment is the adaptation of VGG on CIFAR10 \cite{Liu2015VeryDC}. Results in Fig.~\ref{examples_CIFAR} suggest a relative ranking in performance of studied explainers. We do not include this result in the main manuscript because, to the extend of our knowledge, there is no evaluation on performance of explanations on this dataset. However, we think that this result can serve as a reference for our MNIST and Caltech101 experiments, which is described in Subsection~\ref{overall}.
	
	\onecolumn
	
	\section{Examples of explanations of MNIST, CIFAR10 and Caltech101} \label{exMC}
	
	\begin{figure}[h]
		\centering
		\begin{subfigure}{.2\textwidth}
			\centering
			\includegraphics[scale=0.3]{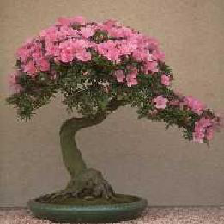}
			\captionsetup{labelformat=empty}
			\caption{Original}
		\end{subfigure}%
		\begin{subfigure}{.2\textwidth}
			\centering
			\includegraphics[scale=0.3]{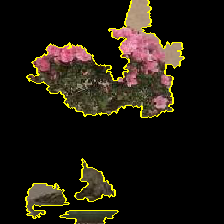}
			\captionsetup{labelformat=empty}
			\caption{1.40}
		\end{subfigure}%
		\begin{subfigure}{.2\textwidth}
			\centering
			\includegraphics[scale=0.3]{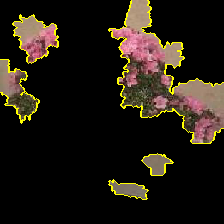}
			\captionsetup{labelformat=empty}
			\caption{1.12}
		\end{subfigure}%
		\begin{subfigure}{.2\textwidth}
			\centering
			\includegraphics[scale=0.3]{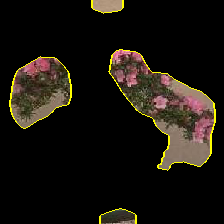}
			\captionsetup{labelformat=empty}
			\caption{1.12}
		\end{subfigure}%
		\begin{subfigure}{.2\textwidth}
			\centering
			\includegraphics[scale=0.3]{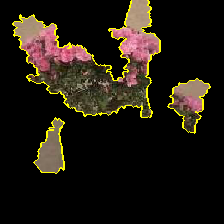}
			\captionsetup{labelformat=empty}
			\caption{1.26}
		\end{subfigure}%
		\\
		\begin{subfigure}{.2\textwidth}
			\centering
			\includegraphics[scale=0.3]{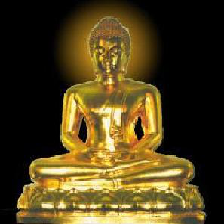}
			\captionsetup{labelformat=empty}
			\caption{Original}
		\end{subfigure}%
		\begin{subfigure}{.2\textwidth}
			\centering
			\includegraphics[scale=0.3]{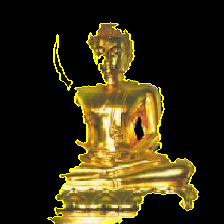}
			\captionsetup{labelformat=empty}
			\caption{2.70}
		\end{subfigure}%
		\begin{subfigure}{.2\textwidth}
			\centering
			\includegraphics[scale=0.3]{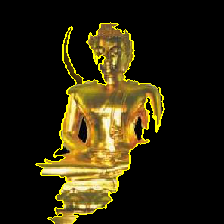}
			\captionsetup{labelformat=empty}
			\caption{2.20}
		\end{subfigure}%
		\begin{subfigure}{.2\textwidth}
			\centering
			\includegraphics[scale=0.3]{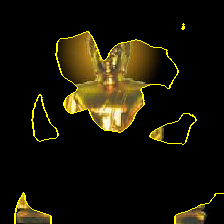}
			\captionsetup{labelformat=empty}
			\caption{1.80}
		\end{subfigure}%
		\begin{subfigure}{.2\textwidth}
			\centering
			\includegraphics[scale=0.3]{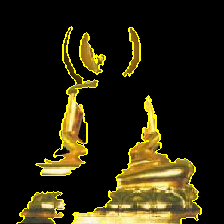}
			\captionsetup{labelformat=empty}
			\caption{1.61}
		\end{subfigure}%
		\\
		\begin{subfigure}{.2\textwidth}
			\centering
			\includegraphics[scale=0.3]{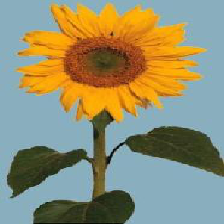}
			\captionsetup{labelformat=empty}
			\caption{Original}
		\end{subfigure}%
		\begin{subfigure}{.2\textwidth}
			\centering
			\includegraphics[scale=0.3]{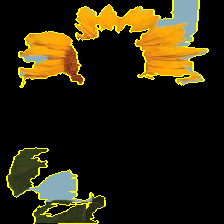}
			\captionsetup{labelformat=empty}
			\caption{1.41}
		\end{subfigure}%
		\begin{subfigure}{.2\textwidth}
			\centering
			\includegraphics[scale=0.3]{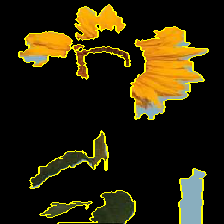}
			\captionsetup{labelformat=empty}
			\caption{1.42}
		\end{subfigure}%
		\begin{subfigure}{.2\textwidth}
			\centering
			\includegraphics[scale=0.3]{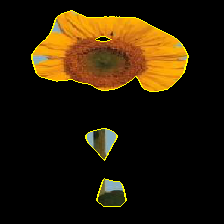}
			\captionsetup{labelformat=empty}
			\caption{1.61}
		\end{subfigure}%
		\begin{subfigure}{.2\textwidth}
			\centering
			\includegraphics[scale=0.3]{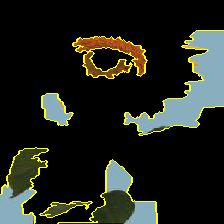}
			\captionsetup{labelformat=empty}
			\caption{1.24}
		\end{subfigure}%
		\\
		\begin{subfigure}{.2\textwidth}
			\centering
			\includegraphics[scale=0.3]{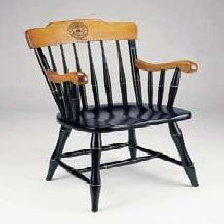}
			\captionsetup{labelformat=empty}
			\caption{Original}
		\end{subfigure}%
		\begin{subfigure}{.2\textwidth}
			\centering
			\includegraphics[scale=0.3]{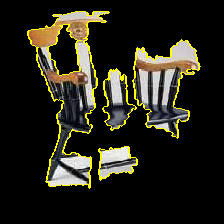}
			\captionsetup{labelformat=empty}
			\caption{3.20}
		\end{subfigure}%
		\begin{subfigure}{.2\textwidth}
			\centering
			\includegraphics[scale=0.3]{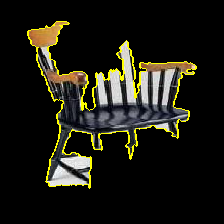}
			\captionsetup{labelformat=empty}
			\caption{2.02}
		\end{subfigure}%
		\begin{subfigure}{.2\textwidth}
			\centering
			\includegraphics[scale=0.3]{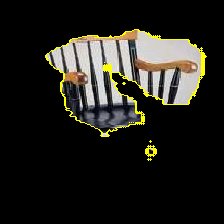}
			\captionsetup{labelformat=empty}
			\caption{4.11}
		\end{subfigure}%
		\begin{subfigure}{.2\textwidth}
			\centering
			\includegraphics[scale=0.3]{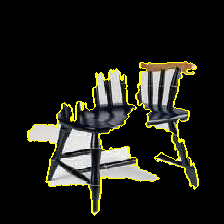}
			\captionsetup{labelformat=empty}
			\caption{2.06}
		\end{subfigure}%
		\\
		\begin{subfigure}{.2\textwidth}
			\centering
			\includegraphics[scale=0.3]{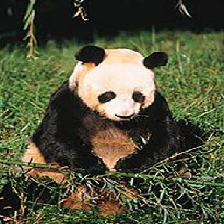}
			\captionsetup{labelformat=empty}
			\caption{Original}
		\end{subfigure}%
		\begin{subfigure}{.2\textwidth}
			\centering
			\includegraphics[scale=0.3]{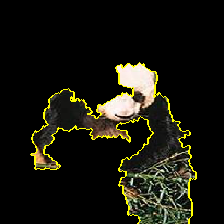}
			\captionsetup{labelformat=empty}
			\caption{1.95}
		\end{subfigure}%
		\begin{subfigure}{.2\textwidth}
			\centering
			\includegraphics[scale=0.3]{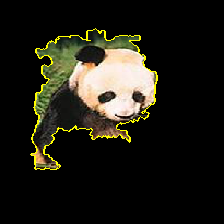}
			\captionsetup{labelformat=empty}
			\caption{2.86}
		\end{subfigure}%
		\begin{subfigure}{.2\textwidth}
			\centering
			\includegraphics[scale=0.3]{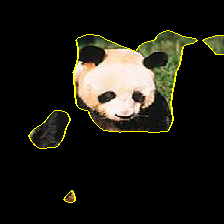}
			\captionsetup{labelformat=empty}
			\caption{2.59}
		\end{subfigure}%
		\begin{subfigure}{.2\textwidth}
			\centering
			\includegraphics[scale=0.3]{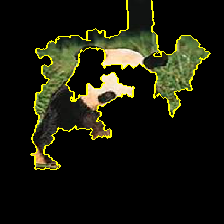}
			\captionsetup{labelformat=empty}
			\caption{1.78}
		\end{subfigure}%
		\\
		\begin{subfigure}{.2\textwidth}
			\centering
			\includegraphics[scale=0.3]{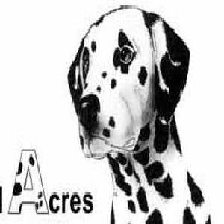}
			\captionsetup{labelformat=empty}
			\caption{Original}
		\end{subfigure}%
		\begin{subfigure}{.2\textwidth}
			\centering
			\includegraphics[scale=0.3]{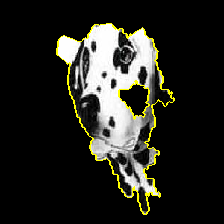}
			\captionsetup{labelformat=empty}
			\caption{3.95}
		\end{subfigure}%
		\begin{subfigure}{.2\textwidth}
			\centering
			\includegraphics[scale=0.3]{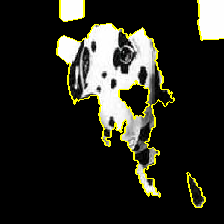}
			\captionsetup{labelformat=empty}
			\caption{1.93}
		\end{subfigure}%
		\begin{subfigure}{.2\textwidth}
			\centering
			\includegraphics[scale=0.3]{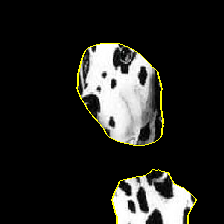}
			\captionsetup{labelformat=empty}
			\caption{3.03}
		\end{subfigure}%
		\begin{subfigure}{.2\textwidth}
			\centering
			\includegraphics[scale=0.3]{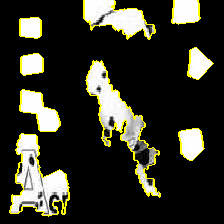}
			\captionsetup{labelformat=empty}
			\caption{1.42}
		\end{subfigure}%
		\\
		\begin{subfigure}{.2\textwidth}
			\centering
			\includegraphics[scale=0.3]{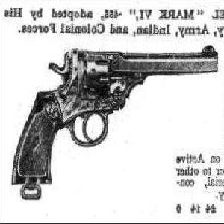}
			\captionsetup{labelformat=empty}
			\caption{Original}
		\end{subfigure}%
		\begin{subfigure}{.2\textwidth}
			\centering
			\includegraphics[scale=0.3]{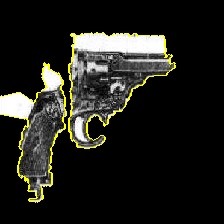}
			\captionsetup{labelformat=empty}
			\caption{9.93}
		\end{subfigure}%
		\begin{subfigure}{.2\textwidth}
			\centering
			\includegraphics[scale=0.3]{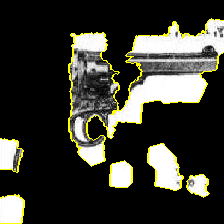}
			\captionsetup{labelformat=empty}
			\caption{3.30}
		\end{subfigure}%
		\begin{subfigure}{.2\textwidth}
			\centering
			\includegraphics[scale=0.3]{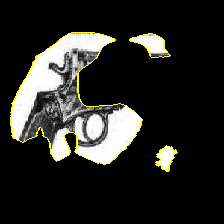}
			\captionsetup{labelformat=empty}
			\caption{3.31}
		\end{subfigure}%
		\begin{subfigure}{.2\textwidth}
			\centering
			\includegraphics[scale=0.3]{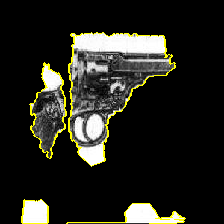}
			\captionsetup{labelformat=empty}
			\caption{8.00}
		\end{subfigure}%

		\caption{Some examples of Explanations and $c$-Eval on Caltech101. The explainers from left to right: SHAP, LIME, GCam and DeepLIFT. The number associated with each figure is the ratio $c_{f, \bm x}(e_{\bm x}) / c_{f, \bm x}(\emptyset)$. We observe that most explanations with high $c$-Eval contain important features of the input images.}
		\label{Caltechapp}
	\end{figure}

\end{document}